%% file: zeus-nsdi23.tex
\documentclass[letterpaper,twocolumn,10pt]{article}
\usepackage{usenix-2020-09}

\usepackage{times}
\usepackage{amsmath,amsthm}
\usepackage{xspace}
\usepackage{graphicx}
\DeclareMathOperator*{\argmin}{argmin}

\definecolor{commentgreen}{rgb}{0, 0.5, 0}
\usepackage[linesnumbered, ]{algorithm2e}
\usepackage{float}

\SetCommentSty{mycommfont}
\newcommand{\algrule}[1][.7pt]{\par\vskip.5\baselineskip\hrule height #1\par\vskip.5\baselineskip}

\usepackage[english]{babel}
\usepackage{subfig}
\usepackage{xcolor}
\usepackage{colortbl}
\usepackage{pifont}
\usepackage[inline]{enumitem}
\usepackage{multirow}
\usepackage{fancybox} 
\usepackage[small, compact]{titlesec}
\usepackage[textfont={it}, font={small,bf}]{caption}
\usepackage[normalem]{ulem}
\usepackage{authblk}
\usepackage{comment}

\def\name{Zeus\xspace}
\def\datasetname{Capriccio\xspace}

\newcommand{\revised}[1]{#1}

\usepackage{tikz}

\newcommand*\blackcircled[1]{\tikz[baseline=(char.base)]{
		\node[shape=circle,draw,fill=black,inner sep=0pt] (char) {\textcolor{white}{#1}};}}

\theoremstyle{definition}

\newtheorem*{defi*}{Definition}

\usepackage{hyperref}
\hypersetup{
	colorlinks=true,      
	linkcolor=blue,       
	citecolor=magenta,    
	filecolor=cyan,       
	urlcolor=red          
}

\usepackage{titling}
\date{}

\usepackage{listings}
\usepackage{courier}
\newenvironment{denseitemize}{
	\begin{itemize}[topsep=2pt, partopsep=0pt, leftmargin=1.5em]
		\setlength{\itemsep}{2pt}
		\setlength{\parskip}{0pt}
		\setlength{\parsep}{0pt}
	}{\end{itemize}}

\newenvironment{denseenum}{
	\begin{enumerate}[topsep=2pt, partopsep=0pt, leftmargin=1.5em]
		\setlength{\itemsep}{2pt}
		\setlength{\parskip}{0pt}
		\setlength{\parsep}{0pt}
	}{\end{enumerate}}

\begin{document}
	
	\date{}
	
	\title{\Large \bf {\name}: Understanding and Optimizing GPU Energy Consumption of DNN Training}
	
	\author{
		\rm Jie You$^*$ \qquad Jae-Won Chung$^*$ \qquad Mosharaf Chowdhury
		\\
		\itshape{University of Michigan}
		} 
	
	

	\pagestyle{plain}
	
  \pagenumbering{gobble} 
	
	
	\maketitle
	
	\input{abstract}
  {\let\thefootnote\relax\footnote{{$^*$Equal contribution.}}}
	\input{intro}
	\input{motivation}
	\input{overview}
	\input{design}
	\input{impl}
	\input{evaluation}
	\input{discussion}
	\input{related}

	\input{outro}
	\input{ack}

	
	\label{EndOfPaper}
	
	\newpage
	{
		\bibliographystyle{plain}
		\bibliography{ref}
	}
	\clearpage

	\input{appendices}

\end{document}

%% file: abstract.tex
\begin{abstract}
Training deep neural networks (DNNs) is becoming increasingly more resource- and energy-intensive every year.
Unfortunately, existing works primarily focus on optimizing DNN training for faster completion, often without considering the impact on energy efficiency.

In this paper, we observe that common practices to improve training performance can often lead to inefficient energy usage.
More importantly, we demonstrate that there is a tradeoff between energy consumption and performance optimization.
To this end, we propose {\name}, an optimization framework to navigate this tradeoff by automatically \revised{finding optimal} job- and GPU-level configurations \revised{for} recurring DNN training jobs.
{\name} uses an online exploration-exploitation approach in conjunction with just-in-time energy profiling, averting the need for expensive offline measurements, while adapting to data drifts over time.
Our evaluation shows that {\name} can improve the energy efficiency of DNN training by \revised{{15.3\%--75.8\%}} for diverse workloads.

\end{abstract}

%% file: intro.tex
\section{Introduction}
\label{sec:intro}

\revised{Deep neural networks (DNNs) have received ubiquitous adoption in recent years across many data-driven application domains such as computer vision~\cite{imagenet,resnet,shufflenetv2}, natural language processing~\cite{bert,albert}, personalized recommendation~\cite{dnn-recommendation-fb,ncf}, and speech recognition~\cite{deepspeech}.} 
\revised{To effectively support such growth, DNN models are predominantly trained in clusters of highly parallel and increasingly more powerful GPUs~\cite{dnn-hardware-survey, rnn-hardware-survey}.}

However, growing demand for computation ultimately translates to greater energy demand.
For instance, training the GPT-3 model~\cite{gpt3} consumes 1,287 megawatt-hour (MWh)~\cite{patterson2021carbon}, which is equivalent to 120 years of electricity consumption for an average U.S. household~\cite{us-household}.
\revised{This trend continues to grow: Meta reports an increasing electricity demand for AI, despite a 28.5\% operational power footprint reduction~\cite{fb-sustainable-ai}.}
Yet, existing literature on DNN training mostly ignores energy efficiency~\cite{greenai}.

We observe that \emph{common performance optimization practices for DNN training can lead to inefficient energy usage}.
\revised{For example, many recent works prescribe large \emph{batch sizes} for higher training throughput~\cite{imagenet1hour,incbs}.
However, we show that maximizing raw throughput may come at the cost of lower energy efficiency.}
Similarly, modern GPUs \revised{allow the configuration of a \emph{power limit} that caps its maximum power draw}, but existing solutions often ignore it.
Our analysis of four generations of NVIDIA GPUs shows that none of them are entirely power proportional, and drawing maximum power \revised{gives diminishing return}.
Indeed, carefully choosing the right batch size and GPU power limit can reduce energy consumption by {23.8\%--74.7\%} for diverse workloads (\S\ref{sec:motivation-eta-gap}).

Unfortunately, reducing energy consumption is not entirely free -- we discover that there is a tradeoff between energy consumption and training time for a given target accuracy (\S\ref{sec:motivation-pareto}).
Our characterization of the \revised{energy-time} Pareto frontier highlights two notable phenomena. 
First, for a given training job, all Pareto-optimal configurations provide varying amounts of energy reductions in comparison to blindly using the maximum batch size and GPU power limit. 
\revised{Second, the amount of reduction in energy consumption often has a non-linear relationship with the increase of training time.}
This raises a simple question: \emph{how do we automatically identify and navigate the tradeoff between energy consumption and training time for DNN training?}

In this paper, we present {\name} to address this question. 
{\name} is a plug-in optimization framework that automatically configures the batch size and GPU power limit to minimize the overall energy consumption and training time for DNN training jobs (\S\ref{sec:overview}). 
Unlike some recent works that only consider GPU-specific configurations \cite{dvfs-impact,dynamic-underclock-gpu}, {\name} simultaneously considers job- and GPU-related configurations. 
\revised{Moreover}, it does not require per-job offline profiling or prediction model training~\cite{gpoeo,odpp}, both of which can be prohibitive in large clusters with heterogeneous hardware and time-varying workloads~\cite{alibaba-trace}.
Instead, {\name} \revised{takes an online exploration-exploitation approach} tailored to the characteristics of DNN training workflows.
\revised{That is, as new data flow into the pipeline, models need to be periodically re-trained~\cite{applied-ml-at-fb}, manifesting itself as \emph{recurring jobs} in production clusters~\cite{applied-ml-at-fb,alibaba-trace}.}
Leveraging this fact, {\name} automatically explores various configurations, measures corresponding gains or losses, and continuously adjusts its actions based on its measurements (\S\ref{sec:design}).

Designing such a solution is challenging due to two sources of uncertainty in DNN training.
First, due to the randomness introduced from DNN parameter initialization and data loading, the energy consumed until a DNN reaches its target accuracy varies even when training is run with the exact same configuration~\cite{randomness-DNN,dawnbench-tta}.
Thus, evaluating a configuration only once does not provide sufficient information about its \emph{expected} energy consumption.
Second, since both DNN models and GPUs have diverse architectures and unique energy characteristics~\cite{benchmark-ai-accelerators}, offline profiling results do not easily generalize to other DNNs and GPUs.
Aggravating these challenges is the large size of the possible configuration space, with each configuration taking hours or even days to evaluate.

{\name} can efficiently determine the optimal set of knobs in the configuration space by \emph{decoupling} the optimization of batch size and power limit without losing optimality.
Specifically, it captures the stochastic nature of DNN training by formulating the batch size optimization problem as a Multi-Armed Bandit (MAB) and runs online optimization under random observations using the Thompson Sampling policy~\cite{thompson1933likelihood}.
Additionally, {\name}'s just-in-time (JIT) energy profiler finds the optimal power limit while training is running, making {\name} a completely online optimization framework.

We have implemented {\name} and integrated it with PyTorch~\cite{pytorch} (\S\ref{sec:impl}).
Evaluation  on a diverse workload consisting of speech recognition, image classification, NLP, and recommendation tasks shows that {\name} reduces energy consumption by \revised{{15.3\%--75.8\%}} and training time by {60.6\%} w.r.t. simply selecting the maximum batch size and maximum GPU power limit.
{\name} converges to optimal configuration among available ones quickly and can adapt to data drift effectively.
{\name}'s benefits expand to multi-GPU settings as well (\S\ref{sec:eval}).

In summary, we make the following contributions:
\begin{denseitemize}
	\item To the best of our knowledge, we are the first to characterize the energy consumption vs. performance tradeoff for DNN training in terms of job- and GPU-specific configuration parameters.
	
	\item We present an online optimization framework that can learn from and adapt to workload dynamics over time.

	\item We implement and evaluate the optimizer in {\name} that integrates with existing DNN training workflows with little code change and negligible overhead, while enabling large benefits.
\end{denseitemize}

\revised{{\name} is open-source and available on GitHub.\footnote{\url{https://github.com/SymbioticLab/Zeus}}}

%% file: motivation.tex
\section{Motivation}
\label{sec:motivation}

In this section, we present an overview of energy consumption characteristics of DNN training on GPUs, opportunities for reducing energy consumption, and conclude with characterizing the tradeoff between reducing energy consumption and improving training performance.

\subsection{DNN Training}
Modern DNNs are trained by going over a large dataset multiple times, where each pass over the dataset is termed an \emph{epoch}~\cite{goodfellowdl}.
One epoch of training consists of thousands of \emph{iterations} of gradient descent over equally sized mini-batches, with the \emph{batch size} affecting model accuracy,\footnote{In this paper, we specifically \revised{consider the} \emph{validation accuracy} of the model, which captures how well the model \revised{generalizes to} unseen data.} training throughput, and energy consumption.
The performance of DNN training is often measured in terms of time-to-accuracy (TTA) for a given target accuracy~\cite{dawnbench-tta}, and increasing training throughput (or precisely goodput~\cite{pollux}) leads to lower TTA.

\revised{Modern DNNs are predominantly trained on increasingly more powerful GPUs}, consuming more energy in the process~\cite{patterson2021carbon,fb-sustainable-ai,treehouse}. 
Recent benchmarks show that GPUs are responsible for around 70\% of the total energy consumption during DNN training~\cite{gpu-70percent,measuring-carbon}.

\revised{In production GPU clusters, as new data flow into the machine learning pipeline, DNNs need to be periodically re-trained at intervals as short as every hour~\cite{applied-ml-at-fb}.}
\revised{This need manifests itself as \emph{recurring jobs} in the GPU cluster~\cite{applied-ml-at-fb,alibaba-trace}.}

\subsection{Opportunities for Improving Energy Efficiency}
\label{sec:motivation-eta-gap}

We \revised{highlight} two job and hardware configurations that can cause sizable energy inefficiency in DNN training: 
(1) \revised{batch size}
and
(2) power limit of the GPU.

\paragraph{Impact of batch size on energy efficiency.}

The size of each mini-batch during DNN training (batch size) determines how many samples are processed in one iteration. 
The higher it is, the faster we can go over the entire input dataset.

\begin{figure}[!t]
	\centering
	\includegraphics[width=0.85\linewidth]{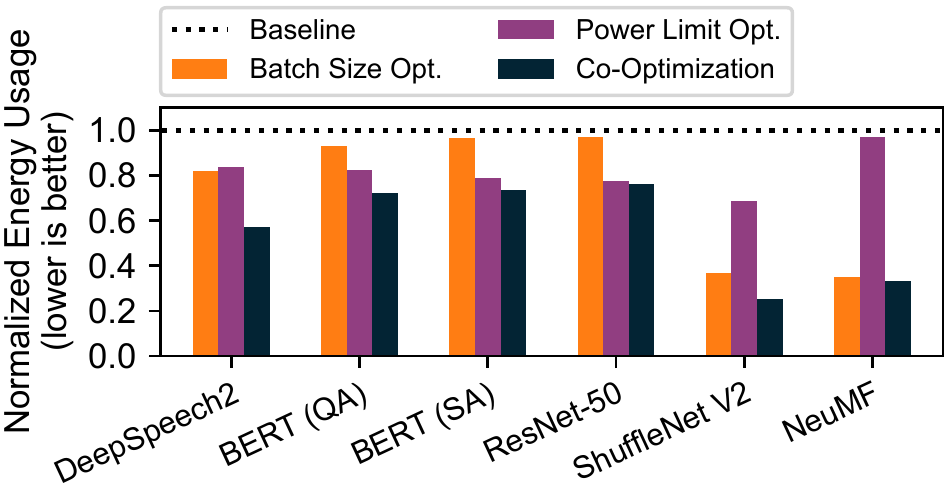}
	\vspace*{-2mm}
  \caption{Energy usage normalized against baseline for DNN training, measured on NVIDIA V100 GPU. Baseline uses maximum power limit and \revised{the default batch size presented in the original model publication when available or the maximum batch size which can consistently reach the target metric.}}
  
	\label{fig:eta-potential}
\end{figure}

\revised{We observe across diverse DNN training workloads that common choices of batch size can lead to more energy consumption for the same target accuracy.}
Specifically, we performed a sweep over a large range of valid batch sizes (from 8 to the maximum batch size that fits in GPU memory)
for six \revised{deep learning workloads including} computer vision (CV), natural language processing (NLP), recommendation, and speech \revised{recognition} on an NVIDIA V100 GPU (Figure~\ref{fig:eta-potential}).\footnote{We measure GPU power consumption \revised{using} NVML~\cite{nvml}.}
Section~\ref{sec:methodology} provides details on workloads and methodology.
We find that the energy-optimal batch size (Batch Size Opt. in Figure~\ref{fig:eta-potential}) can lead to {3.4\%--65.0\%} lower energy consumption than the \revised{default choice} for the same target accuracy.

\begin{figure*}[!t]
	\centering	
	\subfloat[][Energy-Time Tradeoff]{
		\label{fig:eta-tta-with-bound}
		\includegraphics[width=0.382\linewidth]{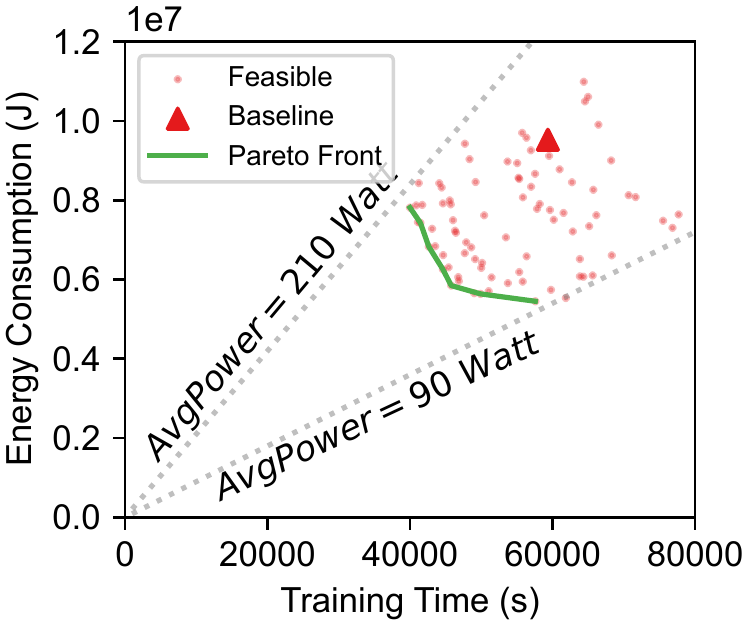}
	}
	\hfil
	\subfloat[][Pareto Front Zoom-in]{
		\label{fig:eta-tta-zoom-in}
		\includegraphics[width=0.382\linewidth]{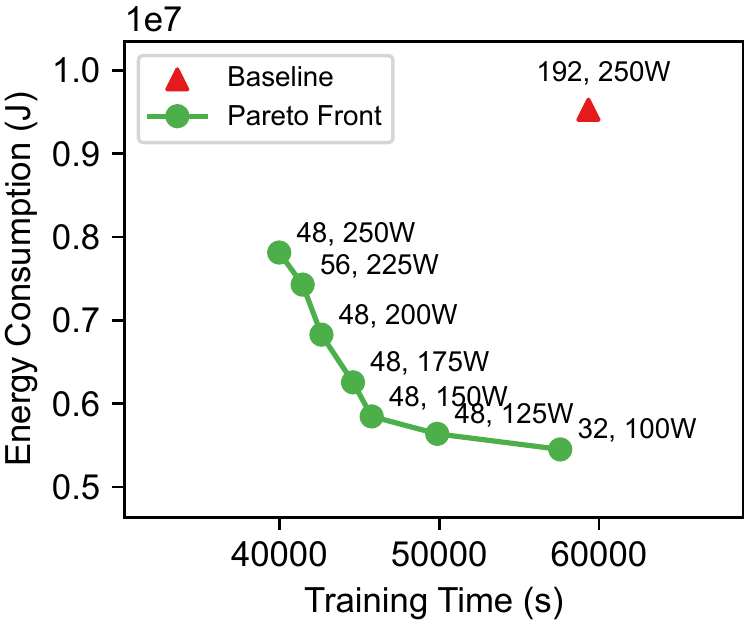}
	}
	
  \caption{DeepSpeech2 trained with LibriSpeech on NVIDIA V100: (a) ETA vs. TTA. The red dots indicate all feasible configurations. The two gray dotted lines indicate two boundaries characterized by average power consumption. The green line indicates the Pareto frontier over all configurations. (b) Zoom-in view on the Pareto frontier \revised{in (a)}, with batch size and power limit annotated on each data point.}
	\label{fig:eta-tta-tradeoff}
\end{figure*}

\paragraph{Impact of GPU power limit on energy efficiency.}
\revised{Setting a GPU's power limit will have the device internally trigger dynamic voltage and frequency scaling (DVFS) such that its power draw does not exceed the power limit~\cite{dvfs-survey}.}
\revised{If not set manually, the power limit is at the maximum by default.}
We performed a sweep over a wide range of GPU power limits\footnote{From the minimum to the maximum power limit allowed by NVIDIA System Management Interface~\cite{nvidia-smi}; from 100W to 250W for NVIDIA V100.} for the aforementioned setup. 
We found that the optimal energy consumption (Power Limit Opt. in Figure~\ref{fig:eta-potential}) may happen at a lower power limit than the maximum and can reduce energy consumption by {3.0\%--31.5\%}. 

\paragraph{Joint optimization.}
As Figure~\ref{fig:eta-potential} shows, we can achieve even more energy savings ({23.8\%--74.7\%} reduction) if we jointly optimize both configurations.
Note that we observed similar opportunities for reducing energy consumption for other generations of GPUs as well (Figure~\ref{fig:appendix-eta-potential} in Appendix~\ref{sec:appendix-eta-potential}). 

\subsection{Energy-Performance Tradeoffs} 
\label{sec:motivation-pareto}

Opportunities for reducing DNN training energy consumption comes with a cost.
When optimized for \revised{energy} efficiency, DNN training performance (\revised{time-to-accuracy, or} TTA) may be impacted.
In the following, we characterize \revised{this} tradeoff.

We define the \revised{energy consumption of DNN training until it reaches its target accuracy} as its \emph{energy-to-accuracy} (ETA):
%
\begin{equation}
	\begin{split}
		\label{eq:eta-tta-power}
		\mathtt{ETA}(b,p) & = \mathtt{TTA}(b,p) \times \mathtt{AvgPower}(b,p) ,
	\end{split}
\end{equation}
where $p$ denotes the GPU power limit, $b$ the batch size, and $\mathtt{AvgPower}(b,p)$ the average power consumption during training with configuration $(b,p)$.
\revised{Similar to TTA, ETA captures the end-to-end goal of DNN training.}

\revised{Note that $\mathtt{AvgPower}(b,p)$ is not the same as the GPU power limit.}
\revised{When changes in configuration $(b,p)$ lead to an increase in TTA, ETA does not always follow because $\mathtt{AvgPower}(b,p)$ can decrease more.}
\revised{This motivates us to investigate the \emph{tradeoff}} between ETA and TTA.

\paragraph{Tradeoff between ETA and TTA.}

We characterize and elaborate on this tradeoff using DeepSpeech2 \revised{trained} on LibriSpeech as an example (Figure~\ref{fig:eta-tta-tradeoff}). 
It shows a scatter plot of (TTA, ETA) for the \revised{batch size and power limit sweep} experiments in Section~\ref{sec:motivation-eta-gap}.
We observe similar results for other workloads as well (Figure~\ref{fig:appendix-eta-tta-pareto} in Appendix~\ref{sec:appendix-eta-tta}).

Let us start with Figure~\ref{fig:eta-tta-with-bound}, where each data point denotes the (TTA, ETA) of training the model for a certain configuration.
While sweeping the configurations, we focus on the boundary of all feasible (TTA, ETA) pairs.
We find them to be bounded by two straight lines characterizing the average GPU power consumption.
\revised{When the GPU is under heavy load, the (TTA, ETA) data points appear closer to 210W.}
\revised{On the other hand, when the GPU is under lighter load, its average power consumption tends closer to 90W, which is close to the GPU's idle power consumption of 70W.}
More importantly, we find a curve along which all (TTA, ETA) pairs achieves Pareto optimality~\cite{censor1977pareto}, for which we cannot improve ETA without sacrificing TTA, and vice versa.

Now let us take a closer look at the Pareto frontier in Figure~\ref{fig:eta-tta-zoom-in}, with the configurations used during training annotated along each data point.
We highlight two takeaways:
\begin{denseenum}
  \item These results \revised{show that} baseline configurations can lead to suboptimal energy efficiency (\S\ref{sec:motivation}). 
		Moreover, it shows that blindly going for high batch size and power limit configurations can lead to suboptimal TTA as well.
	
	\item There exists a tradeoff between ETA and TTA, with different optimums for each. 
		The configuration optimizing the ETA ($b=$32, $p=$100W) is different from that optimizing TTA ($b=$48, $p=$250W). 
	
\end{denseenum}

%% file: overview.tex
\section{{\name} Overview}
\label{sec:overview}

{\name} is an optimization framework that navigates the ETA-TTA tradeoff by automatically configuring the batch size and GPU power limit of recurring DNN training jobs. 
It enables developers to optimize energy and/or performance metrics using a single knob.

\subsection{Optimization Metric} 
\label{sec:hybrid-cost-metric}

Defining a good cost metric for \revised{users} to express their \revised{preference} in this tradeoff is critical in designing {\name}.
We propose a simple cost metric:
\begin{equation}
	\label{eq:cost-metric}
  C(b,p;\revised{\eta}) = \eta \cdot \mathtt{ETA}(b,p)  + (1-\eta) \cdot \mathtt{MAXPOWER} \cdot \mathtt{TTA} (b,p)
\end{equation}

Here $\eta$ is \revised{the parameter specified by the user to express} the relative importance of energy efficiency and training performance (throughput).
When $\eta=0$, we are only optimizing for time consumption, whereas when $\eta=1$, we are only optimizing for energy consumption.
$\mathtt{MAXPOWER}$ is the maximum power limit supported by the GPU, a constant introduced to unify the units of measure in the cost metric.



\subsection{Challenges in Picking the Optimal Configuration}

Combining Equations~\ref{eq:eta-tta-power} and~\ref{eq:cost-metric}, we have:
\begin{equation}
	\label{eq:cost-tta-power}
	C = \left(\eta \cdot \mathtt{AvgPower}(b,p)  + (1-\eta) \cdot \mathtt{MAXPOWER}\right) \cdot \mathtt{TTA} (b,p). 
\end{equation}

Picking the optimal configuration(s) to minimize the energy-time cost $C$ for DNN training is challenging because the search space $[b \times p]$ is large and obtaining the cost of each configuration is difficult. 
\revised{This is because it is hard to determine the value of both $\mathtt{AvgPower}(b,p)$ and $\mathtt{TTA}(b,p)$ efficiently, as explained below.}

\begin{denseitemize}
	\item \textbf{Complex power consumption model:} The total energy consumption of a GPU is affected in a non-linear fashion by both the characteristics of the workload such as the number of instructions and memory accesses, as well as the GPU hardware configurations such as the frequency and voltage of the cores and memory on board~\cite{accelwattch,verified-gpu-instruction-energy-model}.
    \revised{Existing efforts estimate GPU energy consumption based on instruction- or kernel-level information~\cite{integrated-gpu-power-model, gpu-power-model-analytical}, which are architecture-specific and workload-dependent.}

	\item \textbf{Stochastic nature of DNN training:} Modeling and predicting the duration for training a specific model to target accuracy (TTA) is known to be difficult~\cite{tiresias}.
    Moreover, \revised{the} randomness introduced during model initialization and data loading leads to variations of TTA, even when the same job is run on the same GPU with the same configuration -- TTA variations can be as large as 14\%~\cite{dawnbench-tta}.
	
\end{denseitemize}

Fortunately, DNN training jobs often recur in production \revised{clusters}~\cite{applied-ml-at-fb,alibaba-trace}.
This provides opportunities for empirical estimation through repeated measurements across recurrences of the same training job.

\subsection{Architectural Overview}
At a high-level, {\name} takes an online exploration-exploitation approach to minimize the aggregate cost of recurrent DNN training jobs. 
{\name} addresses the aforementioned challenges with two key components:
\begin{denseenum}
  	
  \item A just-in-time (JIT) online profiler, which efficiently profiles the energy characteristics of the training job online. 
	
	\item Multi-Armed Bandit (MAB) with Thompson sampling, which allows us to embrace the stochastic nature of DL training and optimize under uncertainty while also adapting to changing workloads such as data drift.
\end{denseenum}

\revised{The combination of the JIT profiler and MAB makes {\name} a fully online solution, allowing it to immediately begin optimizing for incoming jobs.}

\begin{figure}
	\centering
	\includegraphics[trim=300 110 320 75,clip,width=0.795\linewidth]{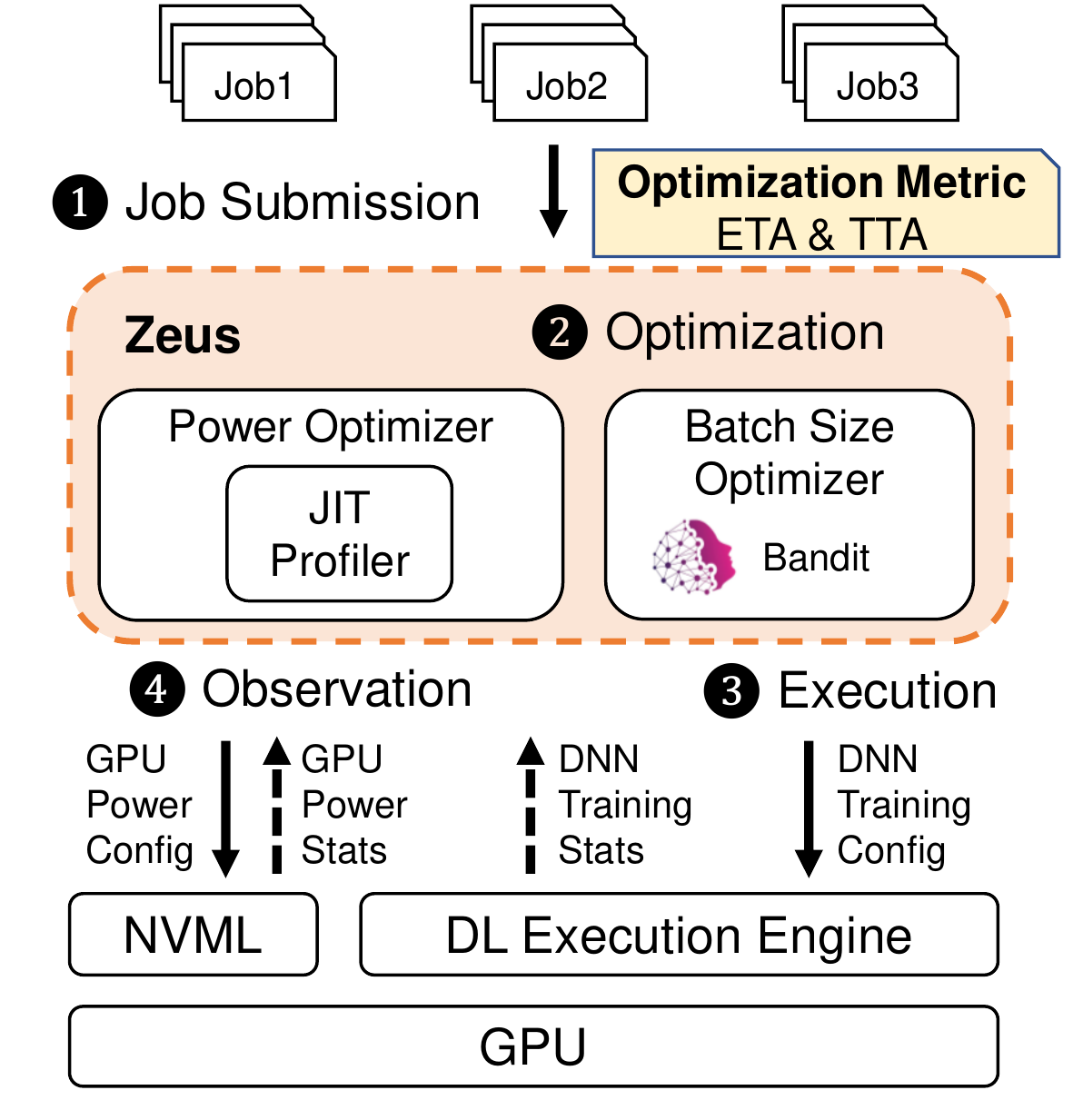}
	\caption{{\name} Workflow.}
	\label{fig:system-overview}
\end{figure}

\paragraph{Workflow of {\name}.}
Figure~\ref{fig:system-overview} shows an overview of the high-level workflow of {\name}.
In a production environment, users submit \blackcircled{1} recurrent DNN training jobs (a tuple of data, model, optimizer, and the target \revised{validation metric}) to {\name}, along with a set of feasible batch sizes $\mathcal{B}$ and power limits $\mathcal{P}$ to explore.
{\name} then predicts \blackcircled{2} the optimal batch size and power limit configuration based on past execution history, and launches \blackcircled{3} the training job with \revised{that} configuration.
During and after the training process, \blackcircled{4} statistics about DNN training (e.g., validation \revised{metric}) and GPU power consumption are collected and fed back to the {\name} optimizer.
The {\name} optimizer learns from the feedback and adjusts its internal states.
The training job will be terminated upon either reaching target \revised{metric} or exceeding a stopping threshold \revised{determined by {\name}}.
The whole process is an automated feedback loop that minimizes the key objective of energy-time cost.

Building {\name} requires both algorithm design and systems support.
Next we describe the core optimization algorithm details (\S\ref{sec:design}) and {\name} implementation highlights (\S\ref{sec:impl}).

%% file: design.tex
\section{{\name} Algorithm Design}
\label{sec:design}

In this section, we delve into the details of how {\name} selects the best batch size and GPU power limit to optimize the overall cost of recurrent DNN training tasks.
We first present the optimization problem formulation and how we decouple the optimizations of batch size and power limit (\S\ref{sec:design-components}).
Next, we show how to optimize power limit (\S\ref{sec:design-plo}) and batch size (\S\ref{sec:design-bso}) under the decoupled framework.
We conclude by discussing how we address common challenging scenarios (\S\ref{sec:extensions}).

\subsection{Problem Formulation}\label{sec:design-components}

The objective of {\name} is to minimize the cost of a recurring job by automatically exploring the feasible set of batch sizes  $\mathcal{B}$ and power limits $\mathcal{P}$.
In essence, we neither want to incur too much cost searching for the optimal configuration, nor do we want to miss it.
Minimizing the \emph{cumulative} cost of the job over recurrences captures the implicit tradeoff between exploration and exploitation.
Put formally in terms of the cost function defined by Equation~\ref{eq:cost-metric}, our objective becomes
\begin{equation}
	\begin{aligned}
		\label{eq:optimization-problem}
		\min_{b,p} \quad    & \sum_{t=1}^T {C(b_t,p_t;\eta)} \\
		\textrm{s.t.} \quad & b_t \in \mathcal{B}, p_t \in \mathcal{P}, \forall t \in [1, T],
	\end{aligned}
\end{equation}
where $b_t$ and $p_t$ \revised{respectively} denote the batch size and power limit chosen at the $t$th recurrence of the job, \revised{and $b$ and $p$ are vectors of length $T$}.

This is a challenging problem without modification, mainly because the size of the search space can be in the order of hundreds, and each value of $C(b, p; \eta)$ inside the search space can only be obtained by running DNN training until it reaches the target metric.
However, further expanding the cost function (Equation~\ref{eq:cost-tta-power}) allows us to \emph{decouple} the exploration of batch size and power limit, making the problem more tractable:
\begin{equation}
	\label{eq:cost-breakdown-epoch}
	\begin{aligned}
		& C(b,p;\eta) \\
	  & =  \left(\eta \cdot \mathtt{AvgPower}(b,p)  + (1-\eta) \cdot \mathtt{MAXPOWER}\right) \cdot \mathtt{TTA} (b,p) \\
		& = \mathtt{Epochs}(b) \cdot \frac{\eta \cdot \mathtt{AvgPower}(b,p) + (1-\eta) \cdot \mathtt{MAXPOWER}}{\mathtt{Throughput}(b,p)}.
	\end{aligned}
\end{equation}
where $\mathtt{Epochs(b)}$ denotes the number of epochs needed to reach the target, and $\mathtt{Throughput(b,p)}$ epochs per second.

We find two key insights that allow the decoupling of batch size $b$ and power limit $p$:
\begin{denseenum}
	\item Given $b$, $\mathtt{AvgPower}(b,p)$ and $\mathtt{Throughput}(b,p)$ can be profiled quickly during training for all possible choices of $p$. 
    This is due to the iterative nature of DNN training, yielding stable power and throughput estimations even with a small number of iterations.
	
	\item $\mathtt{Epochs}(b)$ is not affected by the choice of $p$ as changing the power limit does not change what is computed.
\end{denseenum}

This implies that the optimal power limit, given any batch size, can be determined independently based on online profiling.
Moreover, since any choice of batch size is automatically accompanied by the optimal power limit, our search space is reduced to the set of batch sizes $\mathcal{B}$.

Formally put, we have decoupled the problem in Equation~\ref{eq:optimization-problem} into an equivalent two-level optimization problem
\begin{equation}
	\begin{aligned}
		\label{eq:bso-optimization}
    \min_{b \in \mathcal{B}\revised{^T}} \; & \sum_{t=1}^T {\mathtt{Epochs}(b_t) \cdot \mathtt{EpochCost}(b_t; \eta)} \\
	\end{aligned}
\end{equation}
where 
\begin{equation}
	\begin{aligned}
		\label{eq:plo-optimization}
		& \mathtt{EpochCost}(b_t; \eta) \\
    & = \min_{p_t \in \mathcal{P}} \; \frac{\eta \cdot \mathtt{AvgPower}(b_t,p_t) + (1-\eta) \cdot \mathtt{MAXPOWER}}{\mathtt{Throughput}(b_t,p_t)}. \\
	\end{aligned}
\end{equation}

When a job arrives, {\name} will first decide which batch size to use based on Equation~\ref{eq:bso-optimization} (\S\ref{sec:design-bso}).
Then, based on the batch size, {\name} will pick the optimal power limit based on Equation~\ref{eq:plo-optimization} (\S\ref{sec:design-plo}).

\subsection{Optimizing the Power Limit}\label{sec:design-plo}

We start with how {\name} determines the optimal power limit based on Equation~\ref{eq:plo-optimization}, given a choice of the batch size.
As highlighted earlier, we leverage the iterative nature of DNN training and the recurrent nature of jobs in production DNN training workflows.



When a job with batch size decision $b$ is submitted, our just-in-time (JIT) profiler is triggered and checks if this batch size had been profiled before.
For \revised{an unseen batch size $b$}, it profiles $\mathtt{AvgPower}(b,p)$ and $\mathtt{Throughput}(b,p)$ for all possible power limits $p$ during the first epoch of the job by partitioning the epoch into slices at iteration boundaries and dynamically changing the GPU power limit for each slice.
The profile information is fed back to {\name}, and the optimal power limit of the batch size is determined by solving Equation~\ref{eq:plo-optimization}.
The rest of the epochs are executed with the optimal power limit.
%
Our \emph{online} JIT profiling approach consumes strictly less time and energy compared to offline profiling before running the job, because the profiling process itself contributes to training without affecting its accuracy.
\revised{We show that JIT profiling incurs negligible overhead in Section~\ref{sec:eval-overhead}.}

\subsection{Optimizing the Batch Size}\label{sec:design-bso}

Now we focus on how {\name} determines the batch size $b_t$ for each job recurrence $t$ that optimizes Equation~\ref{eq:bso-optimization}.
As seen in Section~\ref{sec:design-plo}, $\mathtt{EpochCost(b_t;\eta)}$ is a cheap and deterministic function that identifies the optimal power limit for any batch size $b_t$ and returns the optimal cost of one epoch.
Thus, we may limit our exploration to choosing the optimal batch size because whichever batch size we choose, the optimal power limit will accompany it.

Due to the unpredictable and stochastic nature of DNN training, picking out the optimal batch size without adequate exploration is difficult. 
Hence, a good solution must 
(1) incorporate such nature of DNN training into its exploration process, and (2) intelligently tradeoff the cost of exploring for potentially better batch sizes and the gain of exploiting batch sizes that are already known to be good.

\paragraph{Grid search is suboptimal.}
%
We argue that exhaustively going through all batch sizes and selecting the one with the smallest cost is still suboptimal due to the stochastic nature of DNN training.
That is, because the cost of a DNN training job can differ even when executed with the exact same configurations, it must be modeled as a \emph{cost distribution} with unknown mean and variance.
Although performing several trials for each batch size may yield a better estimation of the mean cost, such a strategy leads to \emph{high exploration cost} because it does not quickly rule out obviously suboptimal batch sizes.

\paragraph{Multi-Armed Bandit formulation.}
{\name} aims to explore the cost of different batch sizes and converge to the optimal batch size, while not incurring too much exploration cost.

{\name} formulates the problem as a Multi-Armed Bandit (MAB) with $T$ trials and $B$ arms, where each trial corresponds to a recurrence of the job and each arm to a batch size in $\mathcal{B}$.
MAB is a good fit to our problem scenario in that it \revised{captures the stochasticity of DNN training by modeling the cost of each batch size as a random variable.}
\revised{Specifically, we choose the} Gaussian distribution~\cite{ts-tutorial} due to its representational flexibility.
The objective of the MAB formulation is to minimize the \emph{cumulative cost regret} defined as
\begin{equation}
\label{eq:cumulative-cost-regret}
\sum_{t=1}^T { \mathtt{Regret}(b_t;\eta) }
\end{equation}
where the regret  of choosing $b_t$ is defined as
\begin{equation}
\label{eq:cost-regret}
  \begin{aligned}
  & \mathtt{Regret}(b_t;\eta) \\
  & = \mathtt{Epochs}(b_t) \cdot \mathtt{EpochCost}(b_t;\eta) - \min_{b,p} \; \mathtt{Cost}(b, p; \eta).
  \end{aligned}
\end{equation}
Minimizing cumulative cost regret aligns with our objective in Equation~\ref{eq:bso-optimization}.

\paragraph{Thompson Sampling.}

\begin{algorithm}[t]
\DontPrintSemicolon
\SetNoFillComment
\let\oldnl\nl
\newcommand{\nonl}{\renewcommand{\nl}{\let\nl\oldnl}}
	
	\algrule
	
	\KwIn{Batch sizes $\mathcal{B}$\newline
		Belief posterior parameters $\hat{\mu}_b$ and $\hat{\sigma}_b^2$}
  \KwOut{Batch size to run $b^*$}

  \SetKwFunction{Predict}{Predict}
  \SetKwProg{Function}{Function}{:}{}
    
  \algrule

  \nonl \Function{\Predict{$\mathcal{B}$, $\hat{\mu}_b$, $\hat{\sigma}_b^2$}}{
    \ForEach{\normalfont{batch size} $b \in \mathcal{B}$}{
    \tcc{Sample from the belief distribution}
    Sample $\hat{\theta}_b \sim \mathcal{N}(\hat{\mu}_b, \hat{\sigma}_b^2)$\;
    }

    \tcc{Select the arm with smallest mean cost sample}
    $b^* \gets \argmin_b \hat{\theta}_b$\;
  }

\algrule

\caption{Gaussian Thompson Sampling: Choosing the next batch size to run (\texttt{Predict})}\label{algo:bso-predict}
\end{algorithm}

\begin{algorithm}[t]
\DontPrintSemicolon
\SetNoFillComment
\let\oldnl\nl
\newcommand{\nonl}{\renewcommand{\nl}{\let\nl\oldnl}}

\algrule

  \KwIn{Batch size $b$ and observed cost $C$\newline
        Previous cost observations $C_b$ for $b$\newline
        Belief prior parameters $\hat{\mu}_0$ and $\hat{\sigma}_0^2$}
  \KwOut{Belief posterior parameters $\hat{\mu}_b$ and $\hat{\sigma}_b^2$}

  \SetKwFunction{Observe}{Observe}
  \SetKwProg{Function}{Function}{:}{}
    
\algrule


  \nonl \Function{\Observe{$b$, $C$, $\mathcal{C}_b$, $\hat{\mu}_0$, $\hat{\sigma}_0^2$}}{
    \tcc{Add the most recent cost \revised{observation} to history}
    $\mathcal{C}_b \gets \mathcal{C}_b \cup \{ C \}$ 

    \tcc{Compute the variance of the cost}
    $\tilde{\sigma}^2 \gets Var \left( \mathcal{C}_b \right)$\label{algo:bso-observe-cost-var}

    \tcc{Compute the belief distribution's posterior variance}
    $\hat{\sigma}_b^2 \gets {\left( \frac{1}{\hat{\sigma}_0^2} + \frac{|C_b|}{\tilde{\sigma}^2} \right)}^{-1}$

    \tcc{Compute the belief distribution's posterior mean}
    $\hat{\mu}_b \gets \hat{\sigma}_b^2 \left(\frac{\hat{\mu}_0}{\hat{\sigma}_0^2} + \frac{Sum(\mathcal{C}_b)}{\tilde{\sigma}^2} \right)$
  }

\algrule

\caption{Gaussian Thompson Sampling: Updating the belief distribution (\texttt{Observe})}\label{algo:bso-observe}
\end{algorithm}

We adopt the Thompson Sampling~\cite{ts-tutorial} policy for the MAB formulation to tradeoff exploration and exploitation, not only because it is known to perform well in practice~\cite{empirical-ts,ts-tutorial} and had successful adoption recently~\cite{marcus2021bao, narya}, but also because its modeling assumptions fit our problem scenario well.

At a high level, Thompson Sampling is an online procedure that refines its \emph{belief} about the \emph{mean cost} of each arm (batch size) based on experience.
At each recurrence, the belief is used to pick the arm  with the lowest estimated mean cost (Algorithm~\ref{algo:bso-predict}), and the belief is updated based on the actual cost observed (Algorithm~\ref{algo:bso-observe}).

Specifically, the cost distribution is modeled as a Gaussian distribution with unknown mean $\theta_b$.
Then, the belief about $\theta_b$ is modeled with its conjugate prior distribution, which is also a Gaussian distribution~\cite{conjugate}.
That is, $\theta_b \sim \mathcal{N}(\hat{\mu}_b,\hat{\sigma}_b^2)$.
Here it is important to note that $1/\hat{\sigma}_b^2$ can be thought as of how confident \revised{the policy} is in its belief about that arm, with the confidence increasing as it accumulates more observations of the cost of choosing that arm.
Then, Thompson Sampling automatically balances exploration and exploitation by choosing the arm with the smallest mean cost sample $\hat{\theta}_b \sim \mathcal{N}(\hat{\mu}_b, \hat{\sigma}_b^2)$ (Algorithm~\ref{algo:bso-predict}).
With low confidence (high variance), $\hat{\theta}_b$ will be dispersed across a wider range of costs, having higher chances of getting chosen even if some of its initial observations showed high cost.
In contrast, when the arms observed a lot of cost samples and the confidence is high (low variance), $\hat{\theta}_b$ is likely to be centered around the mean observed cost, allowing the exploitation of arms that are known to be good.
After the actual cost of an arm is observed, the belief parameters of that arm are updated using the Bayes Rule~\cite{ts-tutorial} (Algorithm~\ref{algo:bso-observe}).

The belief prior parameters $\hat{\mu}_0$ and $\hat{\sigma}_0^2$ reflect prior belief about the mean cost of using the batch size for training and the confidence of such belief.
Hence, the choice of prior parameters serve as a way to initialize the arms such that they reflect prior knowledge about the cost of each arm.
If such information is not available, which is our default assumption, it is also possible to initialize the arms with a flat prior that assumes no prior knowledge -- in our case, this is a Gaussian distribution with zero mean and infinite variance.

In contrast to grid search, our formulation using MAB and Thompson Sampling meets the two requirements mentioned earlier.
That is, MAB inherently incorporates the stochastic nature of DNN training in that it models cost as a \revised{random variable}.
Moreover, Thompson Sampling can quickly rule out batch sizes that are obviously suboptimal because the probability of a smaller mean cost being sampled from an arm that observed noticeably large cost is low.

\subsection{Extensions \revised{for Challenging Scenarios}}
\label{sec:extensions}

\paragraph{Handling unknown cost variance.}
Unlike conventional Gaussian Thompson Sampling applications, we may not assume that the variances of the cost of each arm are known.
That is, the cost variance (i.e., how much the cost will fluctuate even when training is run with the same batch size) is not known before any observation.
Moreover, the cost variance depends not only on the batch size, but also on the DNN's robustness to the randomness in parameter initialization and data loading, making it difficult to quantify at the time the MAB is constructed.
Hence, our approach is to \emph{learn} the cost variance as we observe cost samples (Line~\ref{algo:bso-observe-cost-var} in Algorithm~\ref{algo:bso-observe}).

\begin{figure}[!t]
  \centering
  \includegraphics[trim=220 220 320 175,clip,width=0.9\linewidth]{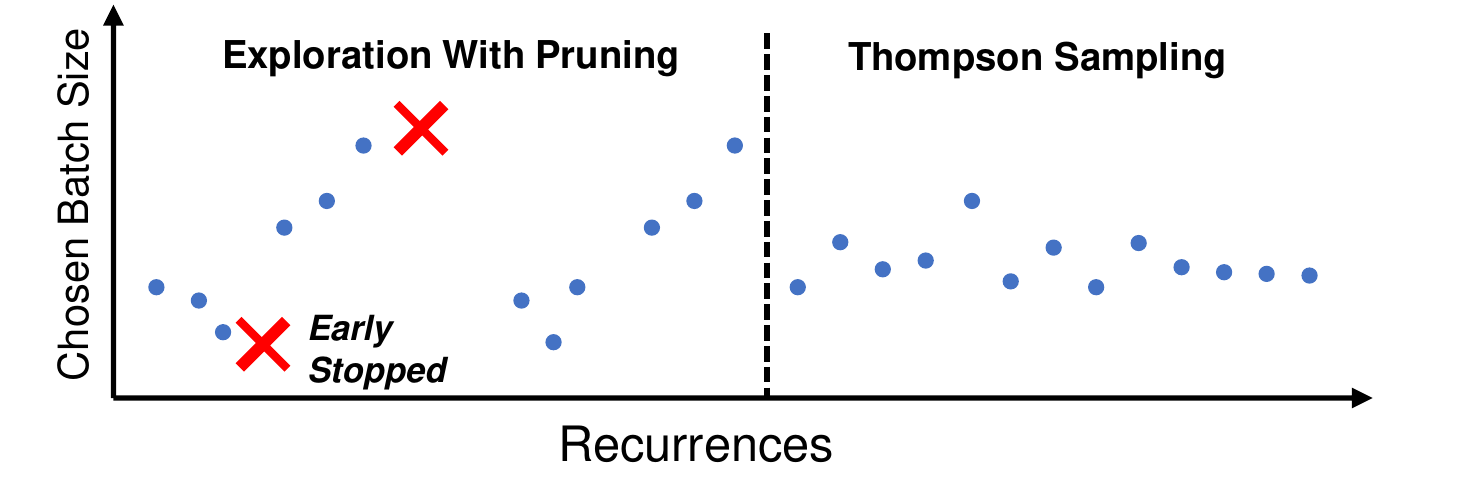}
  \vspace*{-2mm}
  \caption{An example of \revised{batch sizes chosen by {\name} for a recurring job.}
  Each point is a recurrence. \revised{During pruning, {\name} explores each batch size 2 times in order to observe the cost variance (Line~\ref{algo:bso-observe-cost-var} in Algorithm~\ref{algo:bso-observe}).}}
  \label{fig:design-algo-pruning}
\end{figure}

\paragraph{Handling stragglers during exploration.}
There may be cases where an exploratory job does not reach the target metric within a reasonable amount of \revised{cost}, especially during the earlier exploration stage.
To handle this, we employ \emph{early stopping} and \emph{pruning}.
The intuition is that if a batch size does not reach the target metric even after incurring an exceedingly large cost, it is highly unlikely to be the optimal one.

For early stopping, we define a cost threshold $\beta \cdot \min_t C_t$, meaning that when the cost of the current job is to exceed $\beta$ times the minimum cost observed so far, we stop the job and retry with another batch size.
Here $\beta$ is a parameter to account for the stochastic nature of DL training. 
By default, we choose $\beta=2$, with which we should be able to tolerate variations of TTA between different runs of the same configuration, which is usually less than the 14\% \cite{dawnbench-tta}.

\begin{figure}[!t]
  \centering
  \includegraphics[width=0.65\linewidth]{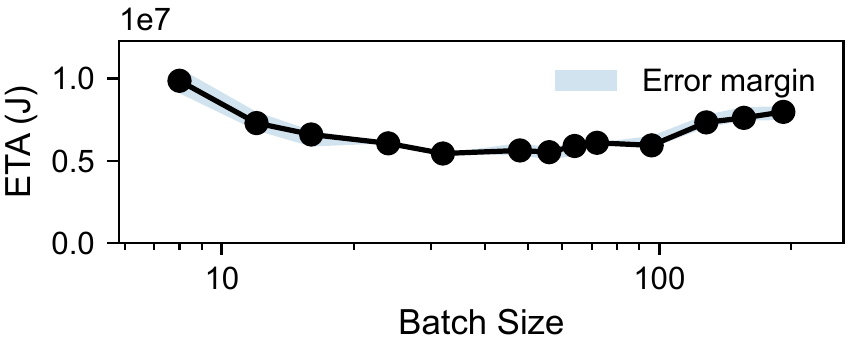}
  \vspace*{-1mm}
  \caption{ETA of each batch size for DeepSpeech2 trained on LibriSpeech. Plots for rest of the workloads are in the Appendix~\ref{sec:appendix-eta-bs-pl}.}
  \label{fig:design-algo-bseta}
\end{figure}

\begin{algorithm}[t]
  \DontPrintSemicolon
  \SetNoFillComment
  \let\oldnl\nl
  \newcommand{\nonl}{\renewcommand{\nl}{\let\nl\oldnl}}
  
  \algrule
  
  \KwIn{Set of batch sizes $\mathcal{B}$\newline
        Default batch size $b_0$\newline
        Belief prior parameters $\hat{\mu}_0$ and $\hat{\sigma}_0^2$}
  
  \algrule
  
  \tcc{\revised{Exploration With Pruning}}
  Recurrence $t \gets 0$\;
  \SetKwFor{RepeatN}{repeat}{times}{end}
  \RepeatN{2}{
    Explore $b_0$\;
    Explore $b < b_0$ until convergence failure\;
    Explore $b > b_0$ until convergence failure\;
    $\mathcal{B} \gets \{b: b~\textnormal{converged}\}$\;
    $b_0 \gets b$ with smallest cost observed\;
    $t \gets t + |\mathcal{B}|$\;
  }
  
  \tcc{Thompson Sampling}
  \While{$t \le T$}{
    $b^* \gets Predict(\mathcal{B}, \hat{\mu}_b, \hat{\sigma}_b^2 \; \forall b \in \mathcal{B})$\;
    Run job with batch size $b^*$ and add cost to $\mathcal{C}_b$\;
    \tcc{Update our belief of the mean cost}
    $\hat{\mu}_b, \hat{\sigma}_b^2 \gets Observe(b, \mathcal{C}_b, \hat{\mu}_0, \hat{\sigma}_0^2)$\;
    $t \gets t + 1$\;
  }
  
  \algrule
  
  \caption{Gaussian Thompson Sampling Batch Size Optimizer.}\label{algo:bso-overall}
\end{algorithm}

For pruning, as illustrated in Figure~\ref{fig:design-algo-pruning}, we begin with the default batch size provided by the user and first try smaller batch sizes until we meet the minimum batch size or a batch size that fails to reach the target metric before the early stopping threshold.
The same process is repeated for batch sizes larger than the default batch size.
Then, only the batch sizes that reached the target metric are kept in the batch size set we explore.
After performing an initial round of pruning, the default batch size is updated to be the one with the smallest cost observed, and we perform pruning once more starting from the new default batch size.

The intuition behind our batch size pruning approach is the convexity we observe in the BS-ETA curve around the optimal batch size (See Figure~\ref{fig:design-algo-bseta}).
Moreover, pruning allows {\name} to quickly rule out batch sizes that are noticeably suboptimal (typically too large, leading to more training epochs and loss of accuracy~\cite{largebatch1,largebatch2}, or too small, yielding gradients that are too noisy~\cite{smallbatch}), thus cutting down the cost of exploration.


The overall process is depicted in Algorithm~\ref{algo:bso-overall}.

\paragraph{Handling concurrent job submissions.}
Classic multi-armed bandit scenarios assume that the MAB immediately observes the cost of pulling an arm.
However, in a DNN training cluster, recurring jobs may overlap in their execution when a later job starts before the completion of an earlier job.
In this case, the MAB does not get to observe the cost of the earlier job at the time it has to decide the batch size for the later job.
For deterministic policies like~\cite{ucb1985,ucb2002}, this leads to duplication exploration of the same batch size back-to-back, reducing the efficiency of exploration.

However, Thompson Sampling naturally mitigates this problem without modification because deciding the next batch size to explore ($\mathtt{Predict}$) is a random function.
That is, because Thompson Sampling \emph{samples} the estimated mean cost from each arm's belief distribution and returns the arm with the lowest sampled value, concurrent jobs \revised{can run} different batch sizes even if there was no information gained between the invocations of $\mathtt{Predict}$.
This is especially the case during the early stage of Thompson Sampling when the arms' belief distributions have large variances (low confidence), losing little exploration efficiency.

During the short initial \revised{pruning} phase, we run concurrent job submissions with the best-known batch size at that time.
As the best batch size constantly updates throughout the exploration stage, this strategy fairly distributes the additional exploration opportunities from concurrent job submissions to batch sizes that are known to converge.
We evaluate {\name}'s efficacy on handling concurrent job submissions in Section~\ref{sec:eval-alibaba}.

\paragraph{Handling data drift.}

In production training clusters, the data on which the model is trained shifts, which is one of the reasons why re-training is triggered~\cite{learning-context-drift,liang2022metashift}.
The implication of drift in the perspective of the MAB is that the cost distribution of each arm is non-stationary.

Thompson Sampling allows a simple modification that allows us to handle non-stationary cost distributions.
Since older cost observations become less and less relevant, we only operate on a window of $N$ most recent cost observations~\cite{nonstationary-bandit}, and the belief distributions will not take old observations into account.
\revised{Unlike exponential decay, windowing also allows the cost variance of the most recent observations to be estimated directly.}
When old history entries are evicted, computing the new parameters of the arm is also cheap thanks to the conjugate prior property. 
This way, {\name} transparently adapts to data drifts in an \emph{online} manner, as we show in Section~\ref{sec:eval-data-drift}.

%% file: impl.tex
\section{\name Implementation}
\label{sec:impl}

{\name} is implemented as a Python library that can be imported into DNN training scripts. 
\revised{The \texttt{ZeusDataLoader} class integrates with PyTorch~\cite{pytorch}.}
The class profiles power and throughput online by slicing epochs in iteration boundaries and invoking the NVML~\cite{nvml} library for power limit configuration and profiling.
We have observed that five seconds of profiling for each power limit is enough to yield stable results.
With the information, the optimal power limit can be automatically determined and applied.
Moreover, \texttt{ZeusDataLoader} monitors the cost incurred by training and early stops the job if needed.
Listing~\ref{lst:design-zeus-integration} shows an example training loop integrated with {\name}.

\begin{lstlisting}[float, language=Python, label=lst:design-zeus-integration, caption={\name} Integration Example]
from zeus import ZeusDataLoader

train_loader = ZeusDataLoader(
    train_set, batch_size, max_epochs, target_metric)
eval_loader = ZeusDataLoader(eval_set, batch_size)

for epoch in train_loader.epochs():  # may early stop
    for batch in train_loader:
        # Learn from batch
    for batch in eval_loader:
        # Evaluate on batch
    train_loader.report_metric(validation_metric)
\end{lstlisting}


\paragraph{Observer Mode.}
\texttt{ZeusDataLoader} supports \emph{Observer Mode}, where it profiles the power consumption and throughput of each power limit and determines the optimal one, but keeps the power limit at the maximum.
By doing so, without affecting time or energy consumption, \texttt{ZeusDataLoader} reports how much time and energy the job \emph{would have} consumed if the power limit were the optimal one, allowing the user to get an idea of the impact of using {\name}.
We believe that such a feature can encourage \name's adoption by informing users of its potential savings.


%% file: evaluation.tex
\section{Evaluation}
\label{sec:eval}

\begin{table*}[]
	\centering
\begin{tabular}{c c c c c c}
	\hline
	\textbf{Task}                 & \textbf{Dataset}     				& \textbf{Model}           					& \textbf{Optimizer} 				& \textbf {$\mathbf{b_0}$ } 		& \textbf{Target Metric}  \\ 
	\hline \hline
	Speech Recognition   & LibriSpeech~\cite{librispeech} & DeepSpeech2~\cite{deepspeech}   & AdamW~\cite{adamw}     			& 192            &    WER = 40.0\%           \\ 
	Question Answering   & SQuAD~\cite{squad-paper}    & BERT (QA)~\cite{bert} 	& AdamW~\cite{adamw}     				& 32             & F1 = 84.0       \\
	Sentiment Analysis   & Sentiment140~\cite{twitter-sentiment} & BERT (SA)~\cite{bert}	& AdamW~\cite{adamw}     			& 128             & Acc. = 84\%       \\
	Image Classification & ImageNet~\cite{imagenet} & ResNet-50~\cite{resnet}   		& Adadelta~\cite{adadelta}  & 256            & Acc. = 65\%   \\
	Image Classification & CIFAR-100~\cite{cifar-paper}   & ShuffleNet-v2~\cite{shufflenetv2} & Adadelta~\cite{adadelta}  	& 1024           	& Acc. = 60\%   \\
	Recommendation       & MovieLens-1M~\cite{movielens}  & NeuMF ~\cite{ncf}            		& Adam~\cite{adam}      			& 1024          &    NDCG = 0.41            \\
	\hline
\end{tabular}
\caption{Models and datasets used in our evaluation. The provided target metrics is the target for each training job. Here $b_0$ denotes the default batch size presented in the original work when feasible, otherwise we choose the maximum batch size which can consistently reach the target. The BERT(QA) and BERT(SA) means fine-tuning BERT on the tasks of question answering and sentiment analysis, respectively.} 
\label{tbl:eval-workloads}
\end{table*}

We evaluate {\name}'s effectiveness in terms of navigating the energy-time tradeoff.
Our key findings are as follows:
\begin{denseenum}
	\item {\name} reduces energy consumption by {15.3\%--75.8\%}. 
		It achieves this by trading off small performance for jobs that are already throughput-optimal; otherwise, it reduces training time by up to {60.1\%} too (\S\ref{sec:eval-performance}). 
	
  \item {\name} \revised{quickly converges to optimal configurations} (\S\ref{sec:eval-performance}).
	
	\item {\name} can handle workloads with data drift (\S\ref{sec:eval-data-drift}) and overall incurs low overhead (\S\ref{sec:eval-overhead}).

  \item {\name} scales to multi-GPU settings (\S\ref{sec:eval-multigpu}) and provides consistent savings across four generations of GPUs (\S\ref{sec:eval-sensitivity}).
  
\end{denseenum}

\subsection{Experimental Setup}
\label{sec:methodology}

\begin{table}[]
	\centering
	\smaller[2]
	\begin{tabular}{l|ll|ll}
		\hline
		\textbf{Node}                                                                                 & \multicolumn{2}{l|}{\textbf{GPU Specification}}           & \multicolumn{2}{l}{\textbf{Host Specification}}               \\ \hline \hline
		\multirow{3}{*}{\begin{tabular}[c]{@{}l@{}}HPE Apollo\\ 6500 Gen10 Plus\\ A40 $\times$ 4\end{tabular}}              & \multicolumn{1}{l|}{Model}  & A40 PCIe  & \multicolumn{1}{l|}{CPU}  & AMD EPYC 7513   \\
		& \multicolumn{1}{l|}{VRAM}   & 48GB      & \multicolumn{1}{l|}{RAM}  & 512GB DDR4-3200 \\
		& \multicolumn{1}{l|}{mArch.} & Ampere    & \multicolumn{1}{l|}{Disk} & 960GB NVMe SSD     \\ \hline
		\multirow{3}{*}{\begin{tabular}[c]{@{}l@{}}CloudLab~\cite{cloudlab}\\ r7525\\ V100 $\times$ 2\end{tabular}}   & \multicolumn{1}{l|}{Model}  & V100 PCIe & \multicolumn{1}{l|}{CPU}  & AMD EPYC 7542   \\
		& \multicolumn{1}{l|}{VRAM}   & 32GB      & \multicolumn{1}{l|}{RAM}  & 512GB DDR4-3200 \\
		& \multicolumn{1}{l|}{mArch.} & Volta     & \multicolumn{1}{l|}{Disk} & 2TB 7200rpm HDD \\ \hline
		\multirow{3}{*}{\begin{tabular}[c]{@{}l@{}}Chameleon\\ Cloud~\cite{chameleoncloud}\\ RTX6000\end{tabular}} & \multicolumn{1}{l|}{Model}  & RTX6000   & \multicolumn{1}{l|}{CPU}  & Xeon Gold 6126  \\
		& \multicolumn{1}{l|}{VRAM}   & 24GB      & \multicolumn{1}{l|}{RAM}  & 192GB           \\
		& \multicolumn{1}{l|}{mArch.} & Turing    & \multicolumn{1}{l|}{Disk} & 256GB SSD       \\ \hline
		\multirow{3}{*}{\begin{tabular}[c]{@{}l@{}}Chameleon\\ Cloud~\cite{chameleoncloud}\\ P100 $\times$ 2\end{tabular}}    & \multicolumn{1}{l|}{Model}  & P100      & \multicolumn{1}{l|}{CPU}  & Xeon E5-2670 v3 \\
		& \multicolumn{1}{l|}{VRAM}   & 16GB      & \multicolumn{1}{l|}{RAM}  & 128GB           \\
		& \multicolumn{1}{l|}{mArch.} & Pascal    & \multicolumn{1}{l|}{Disk} & 1TB HDD      \\
		\hline
	\end{tabular}
  \caption{\revised{Hardware used in the evaluation.}}
	\label{tbl:eval-gpus}
\end{table}

\paragraph{Testbed Setup.}
We evaluate {\name} with four generations of NVIDIA GPUs \revised{as specified in Table~\ref{tbl:eval-gpus}}.

\paragraph{Workloads.}\label{sec:eval-workloads.}


Table~\ref{tbl:eval-workloads} summarizes our workloads.
The default batch size ($b_0$) is chosen from the original model publication when available; otherwise, it is set to be the maximum batch size which consistently achieves the target accuracy.



\revised{In terms of learning rate, models trained with the Adadelta~\cite{adadelta} optimizer do not require an initial learning rate.}
\revised{For optimizers that do require an initial learning rate, we made our best effort in choosing a batch size and learning rate pair that achieves reasonable accuracies by experimenting with values from the original publication of the model and those discovered by popular DL frameworks~\cite{huggingface}.}

\revised{After collecting the initial batch size and learning rate pairs, when we scale the batch size, we applied Square Root Scaling~\cite{lr-scaling-sqr} for adaptive optimizers such as Adam~\cite{adam} following recent theoretical results~\cite{lr-scaling-prescription}.}

\paragraph{Baselines.}

We compare against the following baselines:
\begin{denseenum}
\item \emph{Default ($b=b_0$, $p=\mathtt{MAXPOWER}$).} This is often the default configuration used by practitioners, where the GPU power limit is set to, or rather \emph{not changed from}, the maximum.
	This is the most conservative baseline with no exploration.
	
	\item \emph{Grid Search with Pruning.} This one tries out one configuration of $(b, p)$ for each recurrence of the job and selects the best one.
  We optimize na\"ive grid search by having it prune out batch sizes that failed to reach the target metric. 
	
\end{denseenum}

\paragraph{Metric.}

Our primary metrics are ETA (energy consumption) and TTA (training time). 
Ideally, we want to reduce both; but due to their tradeoff, sometimes it may not be possible to simultaneously do both.



\paragraph{Defaults.}
All experiments are done on NVIDIA V100 GPUs, unless otherwise mentioned.
By default, we highlight $\eta=0.5$ to strike a balance between ETA and TTA.
Later, we sweep $\eta$ from 0 to 1 (\S\ref{sec:eval-sensitivity}).
The early-stopping threshold $\beta$ is set to 2, and we also sweep $\beta$ from 1.5 to 5 (\S\ref{sec:eval-sensitivity}).

\paragraph{\revised{Methodology.}}
\revised{Due to resource constraints and environmental concerns, we cannot afford to repeatedly train all of our workloads with various configurations end-to-end hundreds of times sequentially.}
\revised{However, similar to how {\name} \emph{decouples} the exploration of batch size and power limit, we may apply the same decoupling in our experimentation.}
  \revised{That is, we instead take a trace-driven approach, where we collect two kinds of trace data:}
\begin{denseenum}
\item \revised{Training trace. We train all possible combinations of models and batch sizes until convergence and record the number of epochs the model took to reach its target accuracy. We repeat this with four different random seeds for every combination to capture the stochasticity in DNN training.}

\item \revised{Power trace. We use our JIT profiler to collect the throughput and average power consumption of all possible combinations of model, batch size, and power limit.}
\end{denseenum}

\revised{We then replay these traces when we need to train a model and reconstruct its TTA and ETA values in order to evaluate the decisions made by {\name} and baselines.}
\revised{Moreover, since we have access to all the possible choices and their outcomes, we also know the optimal choice.}
\revised{Therefore, with the traces, we can evaluate the regret achieved by {\name} and baselines.}

\revised{Note that {\name} does not directly learn from these traces (which would be offline-profiling), but instead only learns from the \emph{replay} of these traces in an online fashion.}

\revised{While the aforementioned trace-driven method is used widely throughout our evaluation, we run {\name} end-to-end for the evaluation of handling data drift (\S\ref{sec:eval-data-drift}) because it is more expensive to construct the trace for the drifting dataset.}

\subsection{{\name} Performance}
\label{sec:eval-performance}

In this section, we evaluate the performance of {\name} in terms of energy consumption and training time as well as the convergence characteristics of our Multi-Armed Bandit algorithm.
Each experiment is run across multiple recurrences of DNN training jobs. 
We select the recurrence number to be $2 \cdot |\mathcal{B}| \cdot |\mathcal{P}|$, so that the Grid Search baseline finishes exploration and also has plenty of chances to exploit its choice.


\begin{figure}[!t]
	\centering
	\subfloat[][Energy Consumption]{
		\includegraphics[width=0.86\linewidth]{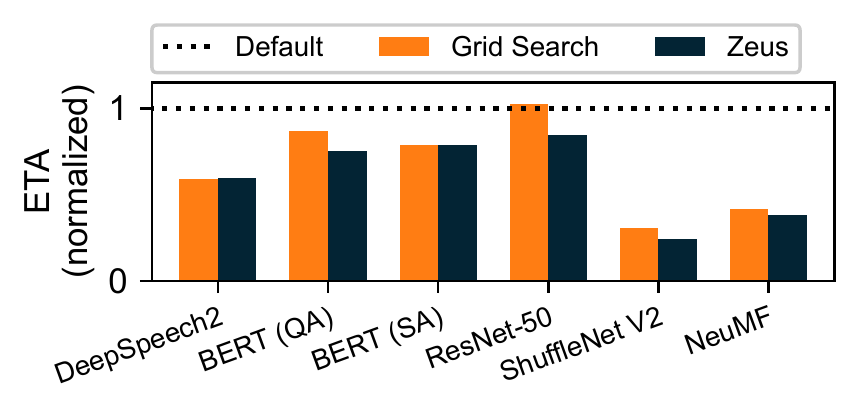}
		\label{fig:eval-cost-foi-energy}
	} \\ 
	\subfloat[][Training Time]{
		\includegraphics[trim=0 0 0 22,clip,width=0.86\linewidth]{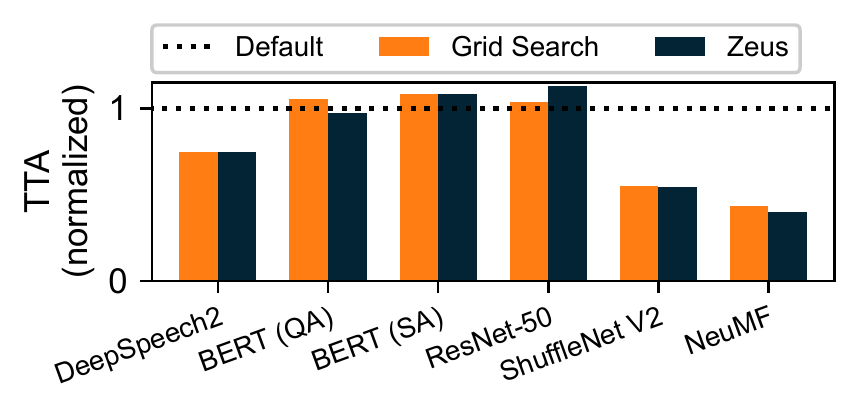}
		\label{fig:eval-cost-foi-time}
	}   
  \caption{{\name} reduces energy consumption for all workloads. (a) energy consumption, (b) training time of each workload, normalized by the Default baseline. Results are computed with the last \revised{five} recurrences, capturing the knobs each method converged to.}
	\label{fig:eval-cost-foi-v100}
\end{figure}

\paragraph{Improvements in ETA.}

Figure~\ref{fig:eval-cost-foi-energy} shows the energy consumption (ETA) of the last \revised{five} recurrences of {\name} and Grid Search w.r.t. the Default baseline, aiming to compare the final point each approach converged to.
{\name} reduces energy consumption (ETA) by up to {15.3\%--75.8\%} w.r.t. the baseline.
This is also comparable to the reduction we found by exhaustively searching through all the configurations in Section~\ref{sec:motivation} as well as by using Grid Search.

\paragraph{Tradeoff with TTA.}
Figure~\ref{fig:eval-cost-foi-time} shows the time consumption (TTA) of the last \revised{five} recurrences of {\name} and Grid Search w.r.t. the Default baseline.
Even though {\name} reduces training time  by up to {60.1\%}, for some workloads TTA is increased by {12.8\%} (Figure~\ref{fig:eval-cost-foi-time}).
This is due to the tradeoff between ETA and TTA, which is the central focus of this paper.
This is especially true for workloads with \revised{a $b_0$ tuned for minimizing training time}, where there is little room for TTA improvement.





\begin{figure}[!t]
	\centering
	\hfil
	\subfloat[][DeepSpeech2]{
		\includegraphics[width=0.45\linewidth]{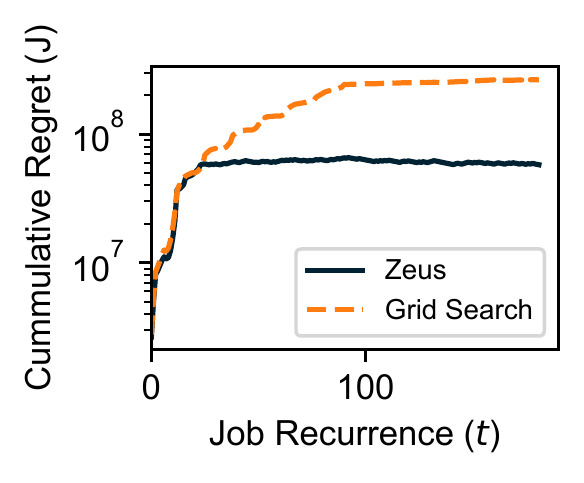}
	}
	\subfloat[][ResNet-50]{
		\includegraphics[width=0.45\linewidth]{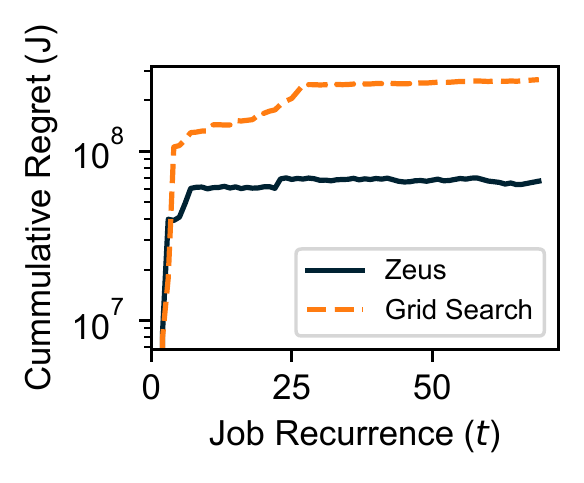}
	}
	\hfil
	\vspace*{-2mm}
	\caption{Cumulative regret of Zeus vs. Grid Search for (a) DeepSpeech2 and (b) ResNet-50. }
	\label{fig:eval-regret}
\end{figure}

\paragraph{Cumulative regret.}
While {\name} and Grid Search perform close to each other, {\name} uses significantly smaller amount of resources to converge.
As a bandit-based solution, the effectiveness of our algorithm can be quantified via regret, the difference between the decision selected and the optimal choice (Equation~\ref{eq:cost-regret} in Section~\ref{sec:design-bso}).

Figure~\ref{fig:eval-regret} shows the cumulative regret of {\name} and Grid Search for DeepSpeech2 and ResNet-50.
The optimal configuration is \revised{identified separately} by an exhaustive parameter sweep.
We observe that in both workloads, {\name} is able to achieve better regret from the first job recurrence.
{\name} reaches the plateau in the cumulative regret earlier than Grid Search, which means it converges to the optimal solution earlier.
We observe similar results for other workload as well (Appendix~\ref{sec:appendix-regret}).
In the worst case, Grid Search \revised{results in $72\times$ more cumulative regret than {\name} until convergence}.

\paragraph{Convergence to a Pareto-optimal configuration.}

\begin{figure}[!t]
	\centering
	
  \subfloat[][{\name}]{
		\includegraphics[width=0.75\linewidth]{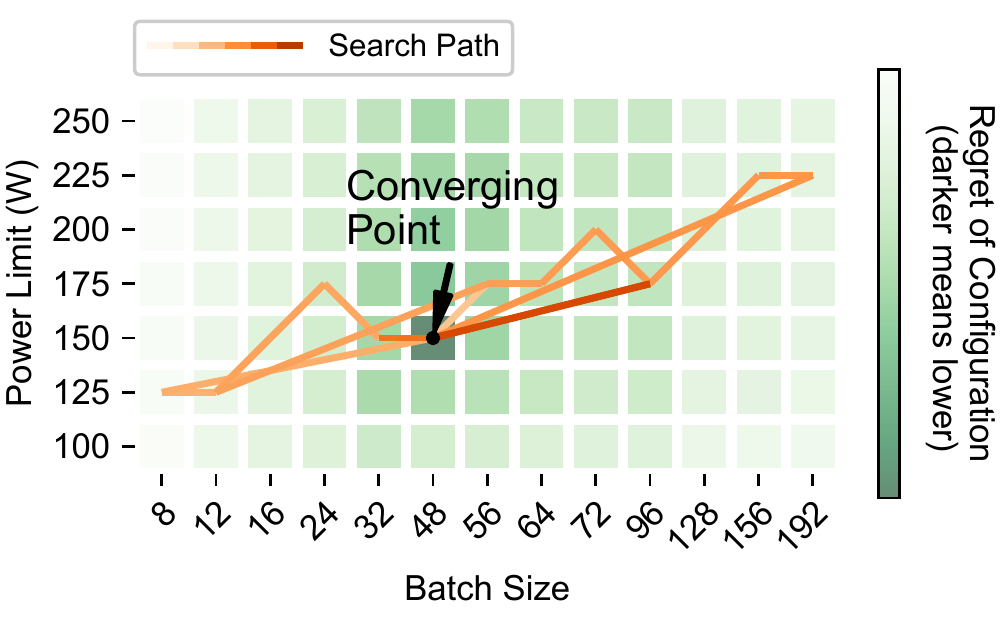}
	}
	
	\subfloat[][Grid Search]{
		\includegraphics[width=0.75\linewidth]{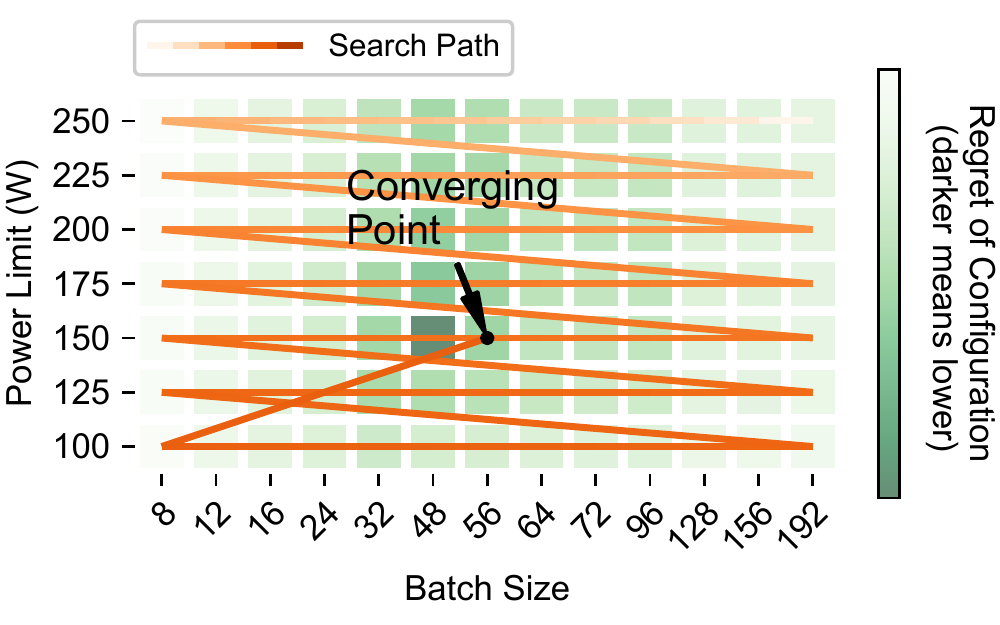}
		\label{fig:eval-covXY-gs}
	} 
	
  \caption{Search paths of (a) {\name} and (b) Grid Search for Deepspeech2. The heatmap in the background shows the regret of each (Batch Size, Power Limit) configuration. Darker background denotes lower regret and therefore better configuration. The colored line with shifting color shows the search path, with darker color being later recurrences.}
  \label{fig:eval-covXY-all}
\end{figure}

Despite having no information about the application beforehand, {\name} learns the energy characteristics of it online in a few iterations.
Figure~\ref{fig:eval-covXY-all} shows the search path of {\name} and Grid Search during training DeepSpeech2.
Due to the decoupling in \revised{the} optimization of power limit and batch size, {\name} explores the configuration space more efficiently and converges to the optimal configuration much faster.
We observe similar results for other workloads (see Appendix~\ref{sec:appendix-searchpath}).
\revised{Moreover}, in Figure~\ref{fig:eval-covXY-gs} we observe that Grid Search may not even converge to optimal configuration.
This is due to the stochastic nature of DNN training, with even the same batch size yielding different energy and time consumptions.
Hence, Grid Search may choose a suboptimal configuration when a suboptimal configuration luckily yields good energy and time consumptions.




\subsection{Trace-Driven Simulation Using the Alibaba Trace}
\label{sec:eval-alibaba}
Here we evaluate how {\name} can save energy and time consumption for DNN training in large clusters. 
We run trace-driven simulation using the Alibaba \revised{GPU cluster} trace~\cite{alibaba-trace} which contains over 1.2 million jobs spanning a period of two months.
The Alibaba GPU cluster trace \revised{is suitable for our evaluation for} two reasons.
First, the trace identifies \revised{groups of} recurring jobs, and each job is annotated with its group ID.
Second, jobs \revised{within} the same group show overlap in their execution, allowing us to evaluate {\name}'s capability of handling concurrent job submissions with Thompson Sampling.

\begin{figure}[!t]
	\centering
	
	\subfloat[][Energy Consumption]{
		\includegraphics[width=0.86\linewidth]{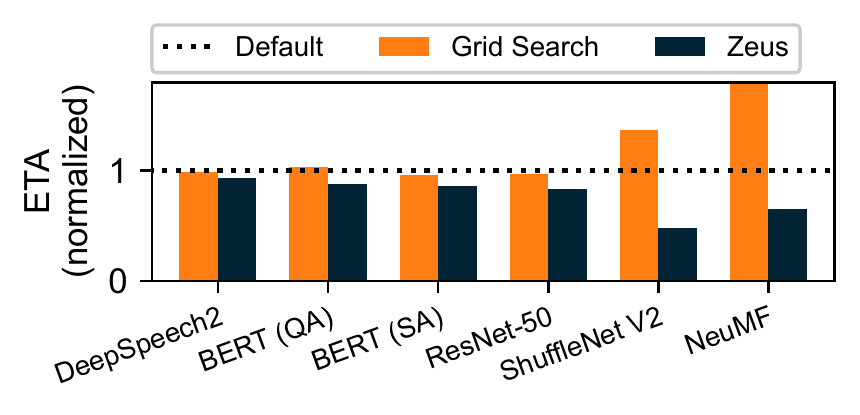}
		\label{fig:eval-alibaba-energy}
	}
	\hfil
	\subfloat[][Training Time]{
		\includegraphics[trim=0 0 0 22,clip,width=0.86\linewidth]{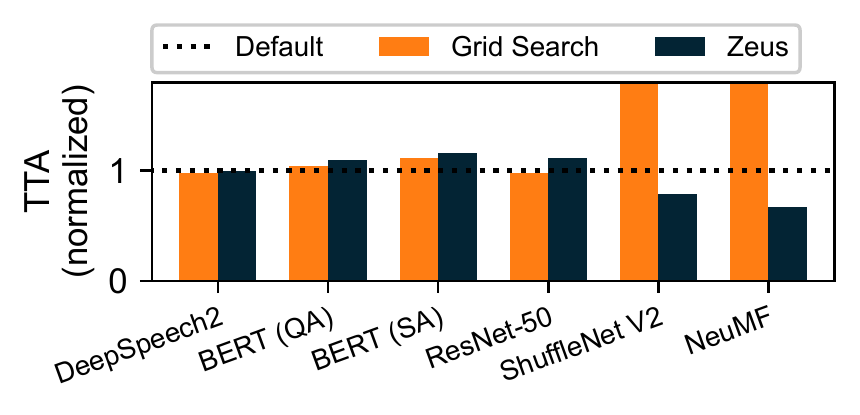}
		\label{fig:eval-alibaba-time}
	}   
	\caption{{\name} reduces energy consumption for all jobs in the Alibaba cluster trace~\cite{alibaba-trace}, compared to Grid Search and Default. (a) Energy consumption with {\name} comparing against baselines, (b) Training time of each type of workload. Both are normalized by the Default baseline. }
	\label{fig:fig:alibaba-all}
\end{figure}

In order to assign job groups to the workload (Table~\ref{tbl:eval-workloads}) that best resembles its runtime, we remove jobs that did not successfully terminate and run K-Means clustering~\cite{kmeans} on the mean job runtime of each group to form six clusters.
Then, we match the six clusters with our six workloads in the order of their mean runtime.
When running simulation, in order to capture the intra-cluster runtime variation of each job, we scale the job runtime with the ratio of the job's original runtime to its cluster's mean runtime.
We compare {\name} with Default and Grid Search and plot the results in Figure~\ref{fig:fig:alibaba-all}.

Figure~\ref{fig:eval-alibaba-energy} shows the cumulative energy consumption of training using all three approaches.
{\name} outperforms both baselines for workloads of all types and sizes.
Note that there are scenarios where the Grid Search performs worse than Default, due to it wasting too much energy and time during the exploration stage.
Thanks to {\name}'s \emph{early stopping} and quick online power optimization, its energy and time cost during the exploration stage is significantly reduced.
Across all the models, {\name} reduces training energy usage by 7\%--52\%.
Figure~\ref{fig:eval-alibaba-time} shows the training time using {\name} to be increased by at most 16\%, and in many cases even decreased by up to 33\%.
Finally, similar to earlier experiments, {\name} had significantly lower cumulative regret than Grid Search.


\subsection{Handling Data Drift}\label{sec:eval-data-drift}

While there are previous works that attempt to identify and address data drift in general ML settings~\cite{learning-context-drift}, existing datasets are classification tasks based on small feature vectors~\cite{electricity-dataset,airlines-dataset}, completely synthetic~\cite{circle-dataset,hyperplane-dataset}, or \revised{private}~\cite{matchmaker}.

Therefore, we create and open-source a new sentiment analysis dataset called \emph{\datasetname} that is suitable for evaluating DNN models.
{\datasetname} consists of 1.6 million tweets over three months from the Sentiment140~\cite{twitter-sentiment} dataset, labeled with sentiment scores and the timestamp of the tweet.
We emulate data drift by capturing a sliding window of 500,000 tweets (roughly the amount of tweets in one month) at a time and moving the window forward by each day, generating 38 slices.
We skip empty dates to avoid having identical slices.

\begin{figure}[!t]
	\centering
	\includegraphics[width=0.8\linewidth]{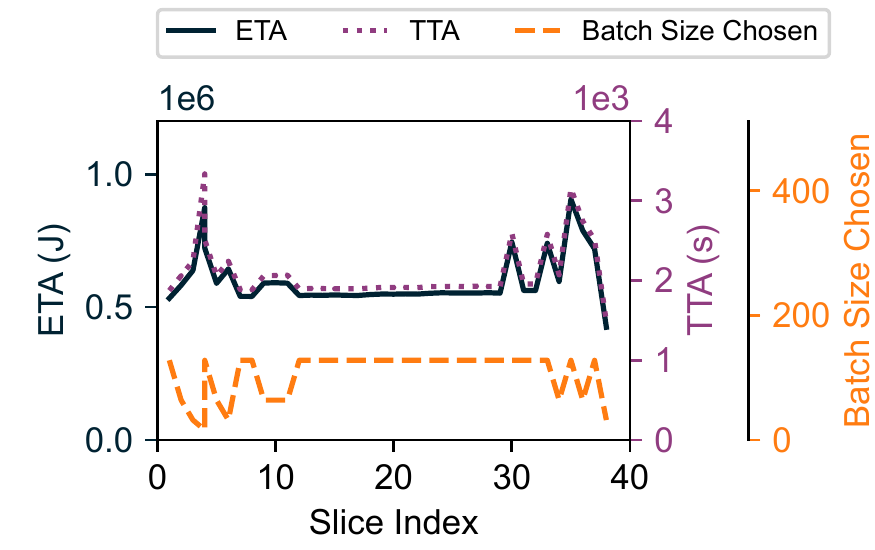}
  \vspace*{-2mm}
  \caption{\revised{Energy and time consumption of training BERT with {\name} on {\datasetname} and the batch size chosen for each slice.}}
	\label{fig:eval-drift-twitter}
\end{figure}

We train BERT~\cite{bert} on {\datasetname} with {\name} configured with a window size of 10, roughly corresponding to a time frame of two weeks on Twitter.
We plot the selected batch size for each recurrence (slice) and its corresponding ETA and TTA of training in Figure~\ref{fig:eval-drift-twitter}.
\revised{It can be seen that spikes in ETA and TTA (signaling that the current batch size may no longer be optimal) trigger the exploration of a batch size that is different from the one previously converged to.}

\subsection{Overhead of JIT Profiling}\label{sec:eval-overhead}


Measurements with the Deepspeech2 model using the default batch size $b_0$ show that JIT profiling results in a 0.01\% increase in energy consumption and a 0.03\% increase in time consumption.
Such a tiny overhead is possible because the time needed to profile all power limits is very small (less than one minute) while one epoch of training spans hours (which is typical for DL workloads).
Measurements on ShuffleNet-v2, which has much shorter epoch duration, show that JIT profiling results in a 0.6\% increase in terms of time consumption and a 2.8\% reduction in energy consumption.





\subsection{Scaling to Multi-GPU}\label{sec:eval-multigpu}

While the primary focus of this paper is on single-GPU settings, in this section, we show that {\name} can be extended to \revised{single-node} multi-GPU training settings by profiling the power consumption of all GPUs that participate in training.
Extensions to distributed multi-GPU \revised{setups that involve network communication} is a potential future work.

Extending to multi-GPU allows us to compare our energy and time consumption with Pollux~\cite{pollux}, a state-of-the-art distributed cluster scheduler that dynamically tunes the batch size \emph{during} DNN training in order to maximize \emph{goodput}.
Training DeepSpeech2 on LibriSpeech on four NVIDIA A40 GPUs, {\name} consumes 12\% more time but 21\% less energy, comparing favorably.
We especially note that while Pollux does not take energy into account, {\name} allows the user to select a different \revised{energy-time} tradeoff point (e.g., speed up training but consume more energy) by selecting an appropriate $\eta$.

\subsection{Sensitivity Analysis and Ablation Studies}
\label{sec:eval-sensitivity}
\paragraph{Impact of $\eta$.}

\begin{figure}[!t]
	\centering
	\includegraphics[width=0.8\linewidth]{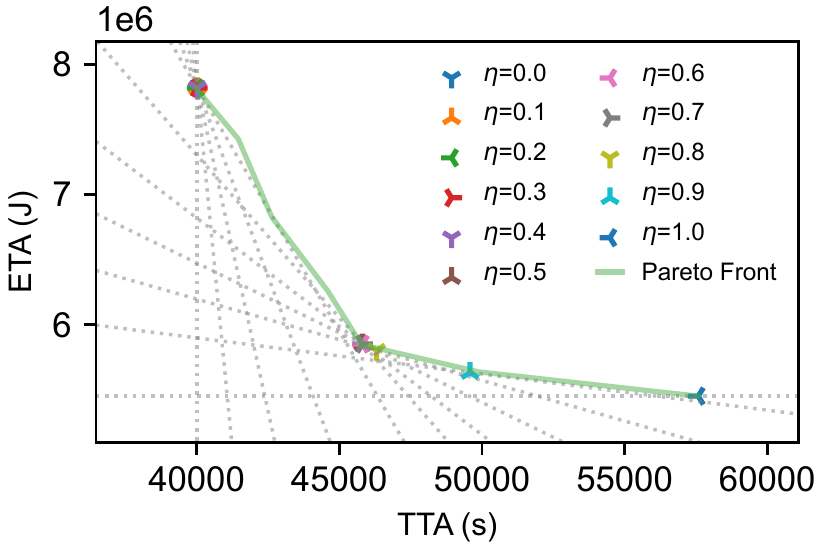}
	\vspace*{-2mm}
  \caption{Pareto Front of DeepSpeech2 and how $\eta$ navigates it.}
	\label{fig:eval-sensitivity-etaknob}
\end{figure}


To characterize the impact of $\eta$ as defined in Equation~\ref{eq:cost-metric}, we perform a sweep of $0 \leq \eta \leq 1$ when training DeepSpeech2 and plot the resulting optimal (TTA, ETA) in Figure~\ref{fig:eval-sensitivity-etaknob}.
We also plot the corresponding Pareto Front for reference.
We observe that the resulting (TTA, ETA) data points fall closely to the Pareto Front.
Moreover, we plot the lines along which the $C$ in Equation~\ref{eq:cost-metric} is a constant, shown as the dotted lines.
As expected, these lines form an envelope around the Pareto Front.
Additional sensitivity analysis for $\eta$ can be found in Appendix~\ref{sec:appendix-sensitivity}.


\paragraph{Impact of early-stopping threshold $\beta$.}

\begin{figure}[!t]
	\centering
	\includegraphics[width=0.65\linewidth]{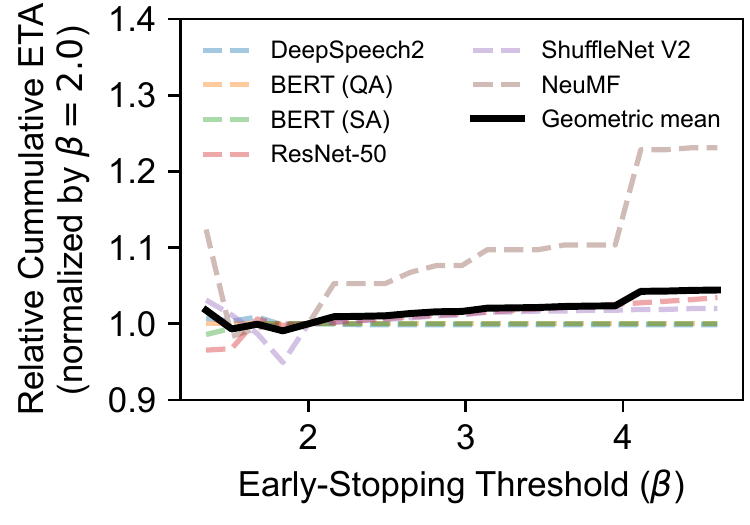}
	\vspace*{-2mm}
	\caption{Relative cumulative energy consumption of {\name} across all jobs, w.r.t. the early-stopping threshold $\beta$.}
	\label{fig:eval-sensitivity-beta}
\end{figure}

To study impact of the early-stopping threshold $\beta$, we sweep $\beta$ from 1.5 to 5 and measure the cumulative ETA across all jobs.
We calculate the difference in ETA relative to our default choice of $\beta=2.0$, and plot the result of all jobs as well as a geometric mean across all jobs in~Figure~\ref{fig:eval-sensitivity-beta}.
The result shows that the default $\beta=2.0$ chosen by {\name} achieves the lowest geometric mean across all jobs.
The intuition behind this is \revised{that when} $\beta$ is too low, {\name} prematurely stops exploratory runs, reducing the effectiveness of exploration.
In contrast, when $\beta$ is too high, it dilutes the benefit of early stopping which leads to inflated \revised{exploration cost}.

\paragraph{Impact of individual components.}

\begin{figure}[!t]
	\begin{minipage}[b]{0.46\linewidth}
	\centering
	\includegraphics[trim=0 6 0 0,clip,width=0.98\linewidth]{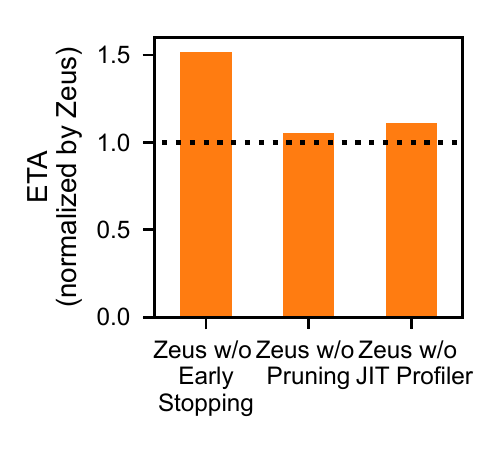}
	\caption{Performance breakdown of {\name}.}
	\label{fig:eval-breakdown-foi-v100}
	\end{minipage}
	\hfil
	\begin{minipage}[b]{0.46\linewidth}
	\centering
	\includegraphics[width=0.98\linewidth]{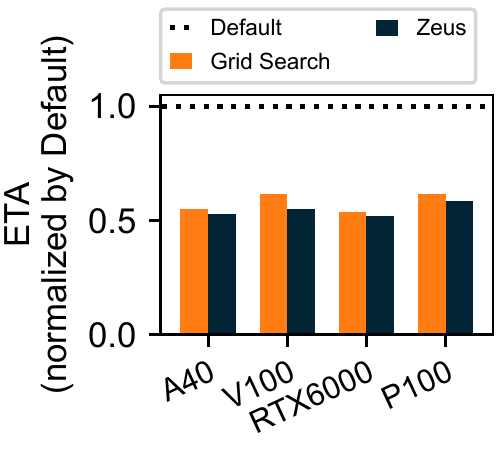}
	\caption{\revised{Normalized ETA w.r.t. GPU models.}}
	\label{fig:eval-sensitivity-gpu}
	\end{minipage}
\end{figure}

In order to show the gains from each component, we show the degradation of removing one component from {\name}: no early stopping (setting $\beta$ to infinity), no pruning (keeping a batch size that didn't reach the target accuracy), and no JIT profiling (profiling each power limit in different recurrences).
%
%
Figure~\ref{fig:eval-breakdown-foi-v100} shows the slowdown relative to {\name} after disabling these components.
We observe that the {\name} benefits mostly from early stopping.

\paragraph{Impact of GPU models.}
	
Figure~\ref{fig:eval-sensitivity-gpu} shows the geometric mean of ETA normalized against Default across all jobs.
\revised{{\name} achieves consistent ETA reductions} across four generations of NVIDIA GPUs.
\revised{See Appendix~\ref{sec:appendix-trace-foi-gpus} for all results.}

%% file: discussion.tex
\section{\revised{Discussion}}
\label{sec:discussion}

\paragraph{Choice of configuration knobs.}
In this paper, we pick the batch size and GPU power limit as the configuration knobs for {\name} to optimize.
We choose these two to strike a balance in the tradeoff between the granularity of control and the size of the search space. 
For instance, one can set the frequency and voltage for individual components on the GPU for more fine-grained control and potentially higher energy efficiency, but this would result in prolonged exploration in the bigger search space.
In contrast, we choose the GPU power limit, which effectively controls both frequency and voltage via DVFS and reduces the search space.

On the DL job configuration side, we pick the batch size as the knob for a similar reason. 
Changing the batch size has a broader impact on energy consumption of end-to-end DNN training, because it affects both the training time and the average power consumption during training.
In comparison, other candidate configuration knobs such as learning rate fall short because they only affect the training time.

\paragraph{Hyperparameter optimization.}
Hyperparameter optimization is an important workload, where many DL training jobs (trials) are submitted with specific hyperparameters chosen from a user-defined search space~\cite{hpo,asha,hyperband,fluid}.
If the users submit these trials with a specific batch size, they can specify the feasible batch size set $\mathcal{B}$ to only contain that single batch size.
In this case, {\name} can still reduce energy consumption by searching for the optimal GPU power limit.

\paragraph{Supporting distributed training.}
{\name} currently only supports single-node training, but it can easily be extended to support distributed scenarios.
Since the same type of GPU will have the same time and power consumption characteristics, we can apply the same power limit configuration across all GPUs to avoid stragglers.
The definition of cost can be extended to sum over the time and energy consumption of all GPUs participating in training, and all other components in our solution can remain identical.

\paragraph{Supporting heterogeneous GPUs.}
Our solution assumes that the training job runs on the same type of GPU across all of its recurrences.
However, in practice, this may not always be possible due to varying resource contention or availability.

It is straightforward to add support for heterogeneous GPUs under our formulation.
That is, cost values observed from one GPU can be \emph{translated} to values that represent the characteristics of another GPU.
As shown in Equation~\ref{eq:bso-optimization}, energy-time cost can be written as the product of $\mathtt{Epochs(b)}$ and $\mathtt{EpochCost}(b;\eta)$.
Here, the former term is independent with the choice of the GPU.
Moreover, the latter term can be quickly profiled on any GPU because it consists of only $\mathtt{AvgPower}(b,p)$ and $\mathtt{Throughput}(b,p)$.
Thus, we can obtain cost values that represent the new GPU by quickly profiling $\mathtt{EpochCost}(b;\eta)$ for each batch size on the new GPU and multiplying it with $\mathtt{Epochs}(b)$ observed from the previous GPU.
These translated cost observations can then be used to learn a new MAB that specializes on the new GPU.

%% file: related.tex
\section{Related Work}

\paragraph{DNN training.}


A large body of recent studies focus on creating fast kernels for tensor operations~\cite{taso,PET,ansor,tvm}, efficiently placing data and/or computation~\cite{pipedream,zero,gspmd,oort}, and optimizing communication~\cite{blinkml,bytescheduler}.
However, most of them optimize for TTA and are oblivious of their energy impact.
These works can be applied together with {\name}, potentially accelerating training while making it energy efficient.


%

Another recent effort in reducing TTA (without considering energy) in multi-GPU DNN training \revised{settings} is Pollux~\cite{pollux}.
Pollux dynamically changes the batch size \emph{during} training based on the Gradient Noise Scale (GNS)~\cite{gns}.
However, GNS does not theoretically capture the generalization of the model~\cite{gns} and can only be efficiently approximated when there are more than one GPUs participating in training.
{\name}, on the other hand, optimizes and trades off TTA and ETA by tuning the batch size \emph{across} job recurrences and does not alter the model's convergence characteristics.


\paragraph{Energy measurement for Deep Learning.}

A recent line of research has analyzed the energy consumption~\cite{patterson2021carbon} as well as the environmental impact~\cite{energy-nlp-policy, quantify-carbon} for training large DNN models inside a cluster.
On the device-level, benchmarking efforts have been made to understand the energy efficiency and performance of training DNN on GPUs and other accelerators~\cite{benchmark-ai-accelerators}.
Several Python frameworks have been built for measurement~\cite{experiment-impact-tracker, build-ontopof-impact-tracker} and prediction~\cite{carbontracker} of energy consumption for DNN training. 
{\name} takes a similar software-based approach to measure power consumption via NVML~\cite{nvml}, in order to perform JIT profiling of DNN training jobs.

\paragraph{Energy optimization for Deep Learning.}

Existing work has investigated energy-accuracy tradeoff in the context of DNN inference with new neural network architecture~\cite{alert} and algorithm-hardware co-design~\cite{edgebert}, and training strategies such as warm-start~\cite{warmstart-training} and gradient-matching-based data subset selection~\cite{gradmatch-icml}. 
Other works optimize energy for DNN training on multiple GPUs with scheduling~\cite{energy-clusterman} and task mapping~\cite{dvfs-taskmap-powercap}.
{\name} complements these solutions as it can be plugged in transparently into these frameworks.

Several works have studied the impact of GPU \revised{dynamic frequency and voltage scaling (DVFS)} and power configuration on the energy consumption and performance of DNN training~\cite{odpp, gpoeo, dvfs-impact, dynamic-underclock-gpu,dvfs-taskmap-powercap}, wherein they focus on the tradeoff between the transient metric of system throughput and power consumption.
\revised{While these work rely on offline modeling and profiling, {\name} focuses on a more realistic end-to-end metric of energy-to-accuracy and is fully online.}

BatchSizer~\cite{batchsizer} introduces batch size as a control knob to optimize for energy efficiency of DNN inference.
{\name} focuses on DNN training, and takes a holistic approach, optimizing both GPU and job configurations together.


%



%% file: outro.tex
\section{Conclusion}

In this work, we sought to understand and optimize the energy consumption of DNN training on GPUs.
We identified the tradeoff between energy consumption and training time, and demonstrated that common practices can lead to inefficient energy usage.
{\name} is an online optimization framework for recurring DNN training jobs that \emph{finds} the Pareto frontier and allows users to \emph{navigate} the frontier by automatically tuning the batch size and GPU power limit of their jobs.
{\name} outperforms the state-of-the-art in terms of energy usage for diverse workloads and real cluster traces by continuously adapting to dynamic workload changes such as data drift.
We earnestly hope that {\name} will inspire the community to consider energy as a first-class resource in DNN optimization.

%% file: ack.tex
\section*{\revised{Acknowledgements}}

\looseness=-1
\revised{Special thanks to CloudLab and Chameleon Cloud for making {\name} experiments possible. We would also like to thank the reviewers, our shepherd Jayashree Mohan, and SymbioticLab members for their insightful feedback. We also thank our colleague Rui Liu for his helpful suggestions. This work is in part supported by NSF grants CNS-1909067 and CNS-2104243 and a grant from VMWare. Jae-Won Chung is additionally supported by the Kwanjeong Educational Foundation.}

%% file: appendices.tex
\appendix

%
%

\section{Energy Savings Potential on GPUs}
\label{sec:appendix-eta-potential}

\begin{figure}[!h]
	\centering

		\subfloat[][NVIDIA A40.]{
				\includegraphics[width=0.7\linewidth]{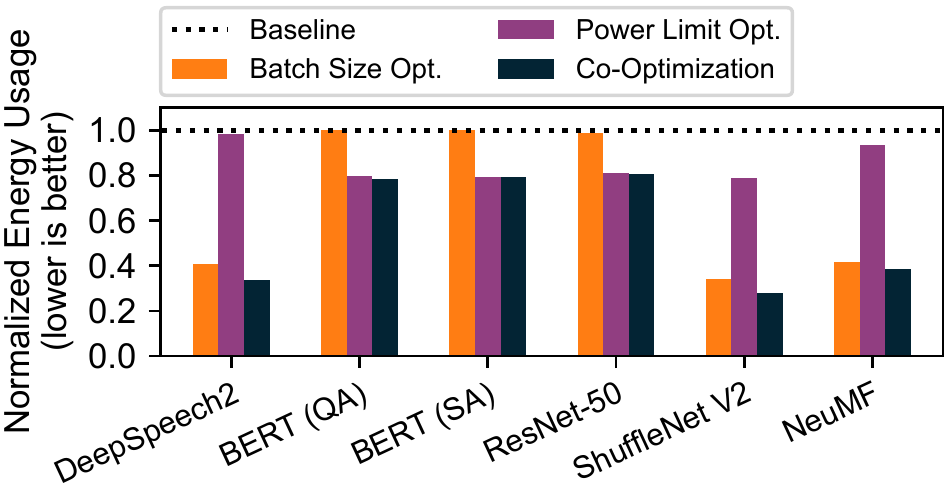}
			}
		\hfil
		\subfloat[][NVIDIA V100.]{
				\includegraphics[width=0.7\linewidth]{Figures/motivation/eta-potential-all-v100.pdf}
			}
		\hfil
		\subfloat[][NVIDIA RTX6000.]{
				\includegraphics[width=0.7\linewidth]{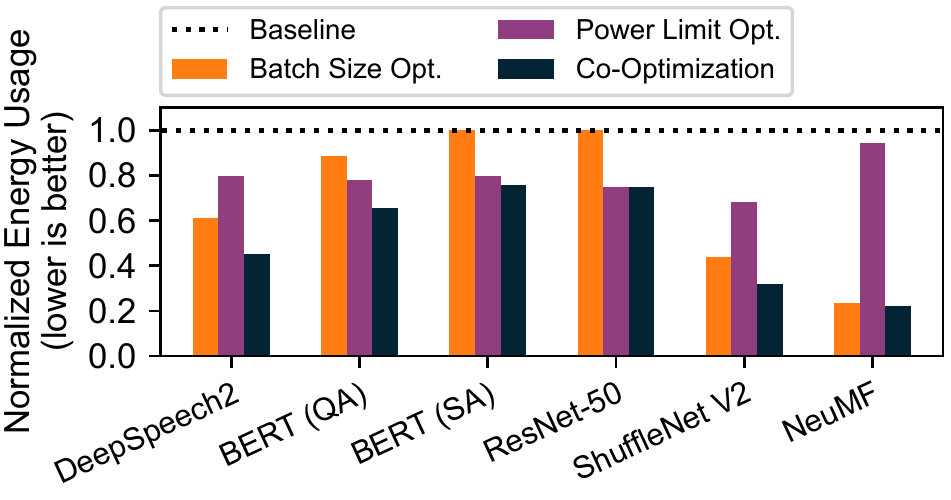}
			}
		\hfil
		\subfloat[][NVIDIA P100.]{
				\includegraphics[width=0.7\linewidth]{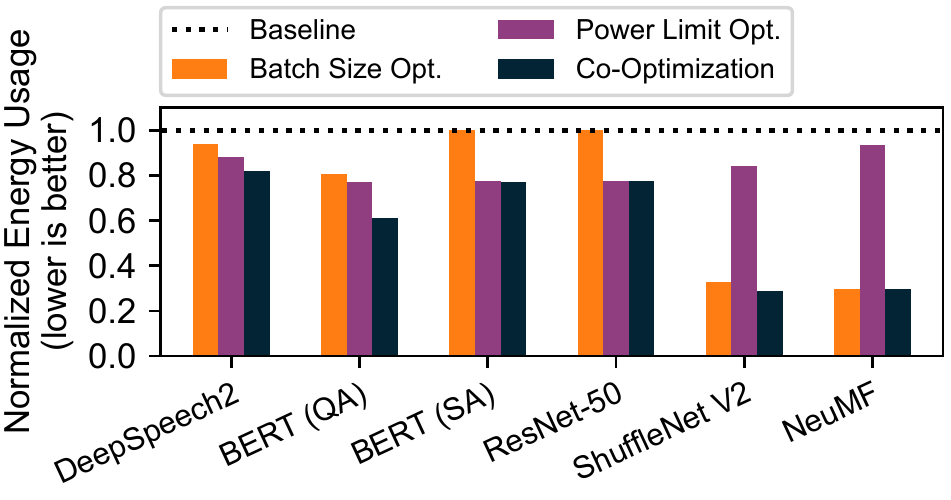}
			}

	\caption{Energy usage normalized against Baseline for DNN training, measured on (a) NVIDIA A40 GPU, (b) NVIDIA V100 GPU, (c) NVIDIA RTX6000 GPU and (d) NVIDIA P100 GPU.}
	\label{fig:appendix-eta-potential}
\end{figure}

Figure~\ref{fig:appendix-eta-potential} shows the potential for energy savings on four different generations of NVIDIA GPUs: Ampere (A40), Volta (V100), Turing (RTX6000), and Pascal (P100).
All four generations show that there are sufficient potential for energy savings, motivating {\name}.

\section{TTA vs. ETA for All Workloads}\label{sec:appendix-eta-tta}

\begin{figure}[!t]
	\centering
	\hfil
	\subfloat[][DeepSpeech2]{
		\includegraphics[width=0.45\linewidth]{Figures/model/pareto-librispeech.pdf}
	}
	\hfil
	\subfloat[][BERT (QA)]{
		\includegraphics[width=0.45\linewidth]{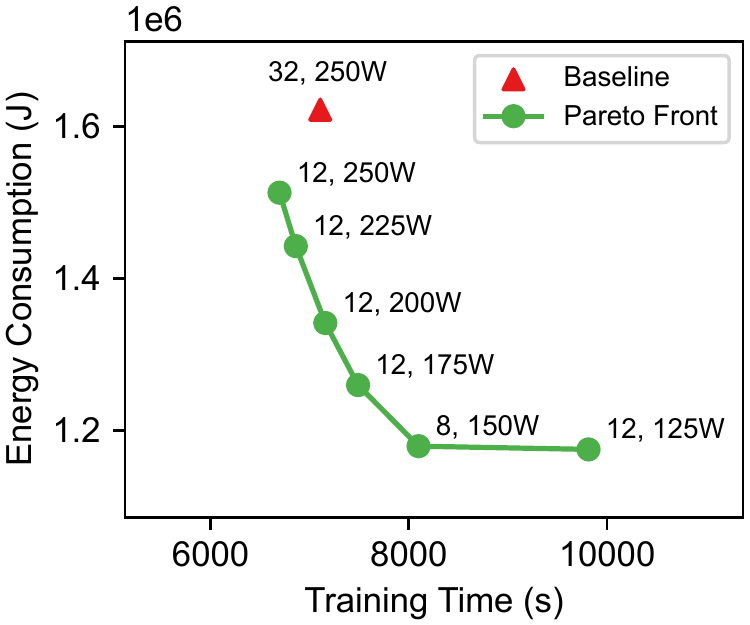} 
	}
	\hfil
	\subfloat[][BERT (SA)]{
		\includegraphics[width=0.45\linewidth]{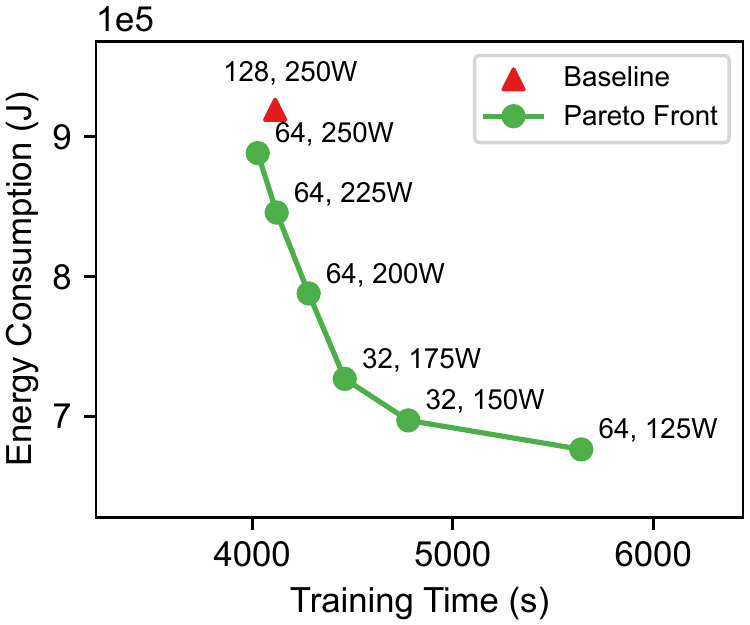}
	}
	\hfil
	\subfloat[][ResNet-50]{
		\includegraphics[width=0.45\linewidth]{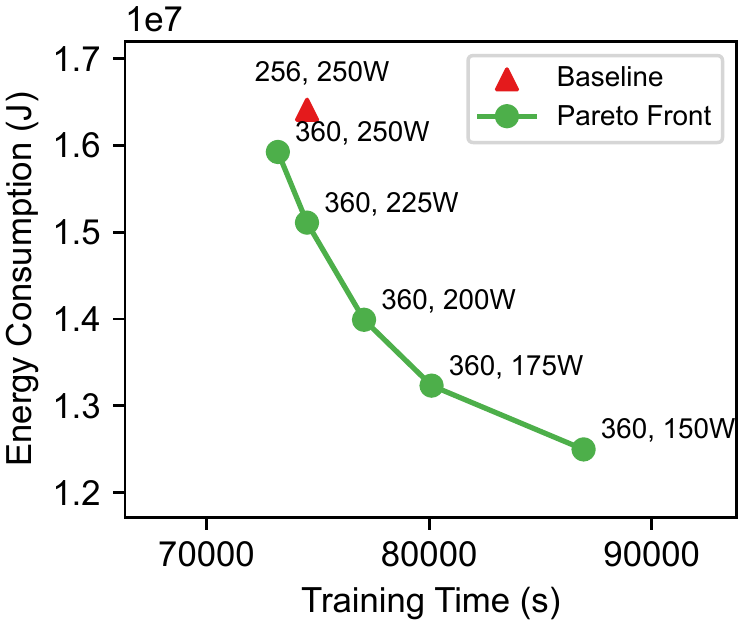}
	}
	\hfil
	\subfloat[][ShuffleNet V2]{
		\includegraphics[width=0.45\linewidth]{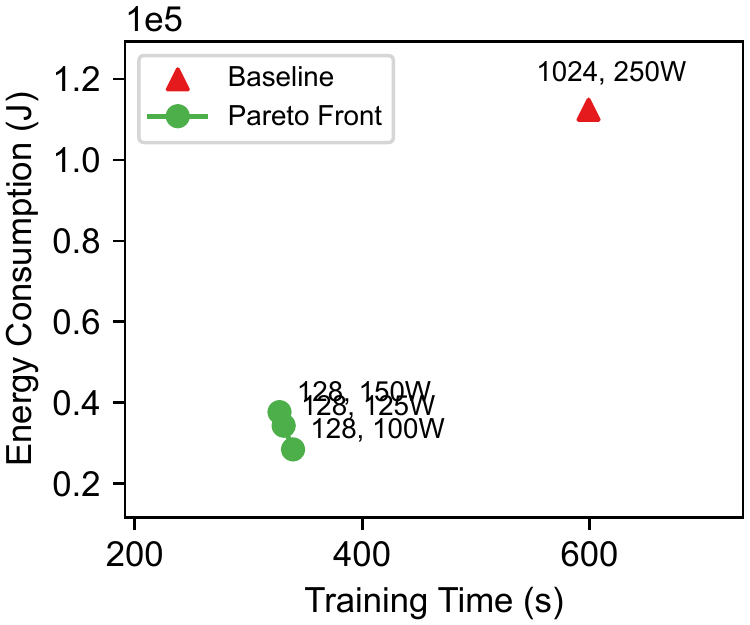}
	}
	\hfil
	\subfloat[][NeuMF]{
		\includegraphics[width=0.45\linewidth]{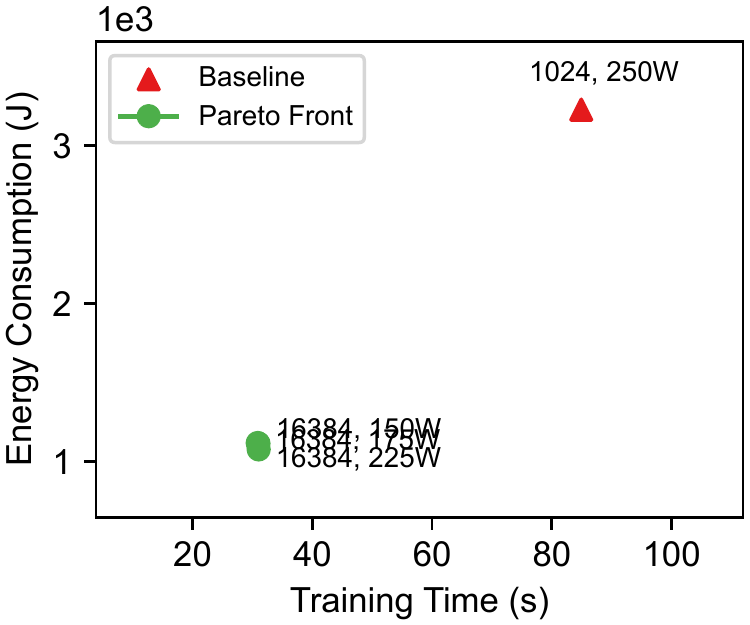}
	}
	\hfil
	\caption{ETA vs. TTA across all workloads, with Pareto Front and default configuration highlighted. Measured on an NVIDIA V100 GPU.}
	\label{fig:appendix-eta-tta-pareto}
\end{figure}

Figure~\ref{fig:appendix-eta-tta-pareto} plots the Pareto Front for all six workloads and the baseline (default batch size and maximum power limit) is shown as a red triangle.
Note that the axes do not start from zero in order to zoom into the Pareto Front.
Data points were gathered on an NVIDIA V100 GPU.

\section{ETA w.r.t. Configurations for All Workloads}\label{sec:appendix-eta-bs-pl}

\begin{figure}[!t]
	\centering
	\hfil
	\subfloat[][DeepSpeech2]{
		\includegraphics[width=0.45\linewidth]{Figures/model/bs-eta-librispeech.pdf}
	}
	\hfil
	\subfloat[][BERT (QA)]{
		\includegraphics[width=0.45\linewidth]{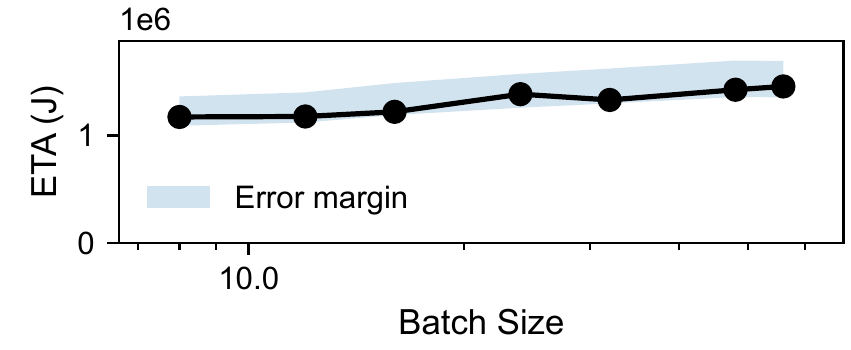} 
	}
	\hfil
	\subfloat[][BERT (SA)]{
		\includegraphics[width=0.45\linewidth]{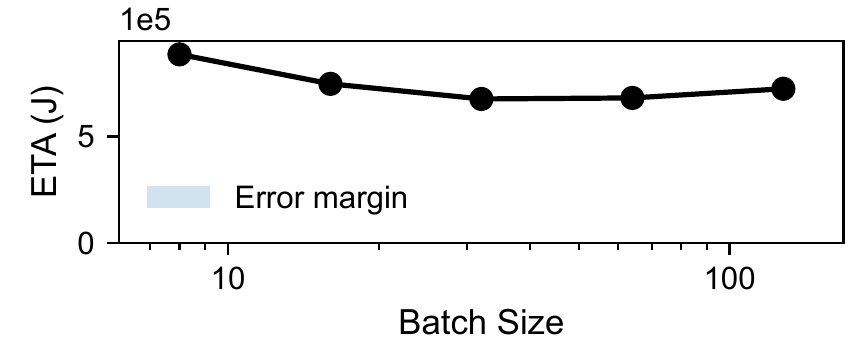}
	}
	\hfil
	\subfloat[][ResNet-50]{
		\includegraphics[width=0.45\linewidth]{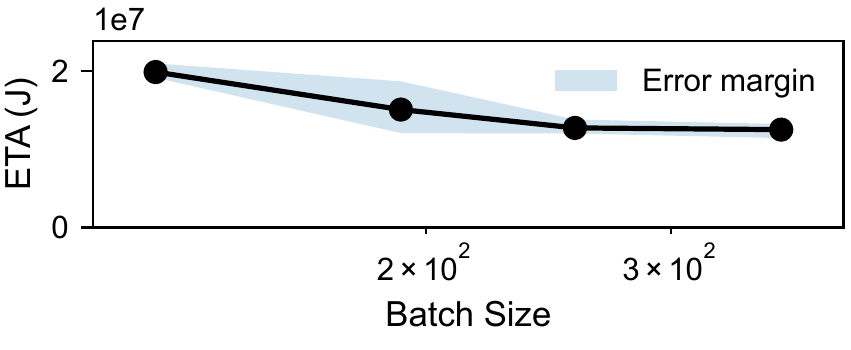}
	}
	\hfil
	\subfloat[][ShuffleNet V2]{
		\includegraphics[width=0.45\linewidth]{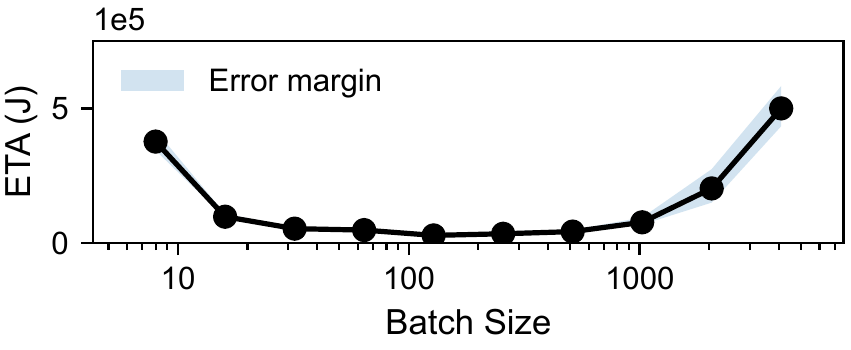}
	}
	\hfil
	\subfloat[][NeuMF]{
		\includegraphics[width=0.45\linewidth]{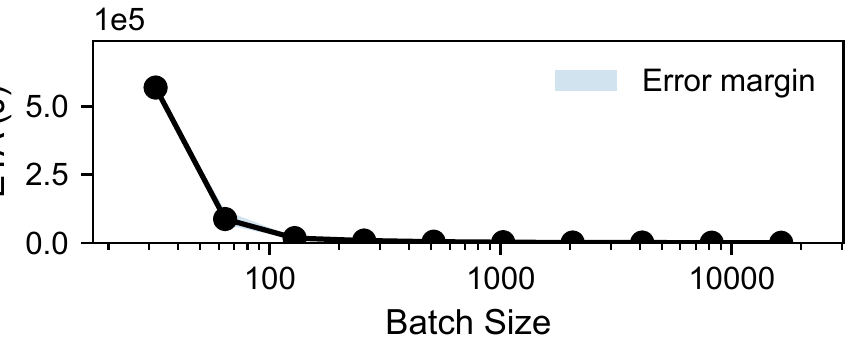}
	}
	\hfil
	\caption{ETA w.r.t batch size of different DNN training workload. The blue shade shows the error margin across repeated runs.}
	\label{fig:appendix-bs-eta}
\end{figure}

\begin{figure}[!t]
	\centering
	\hfil
	\subfloat[][DeepSpeech2]{
		\includegraphics[width=0.45\linewidth]{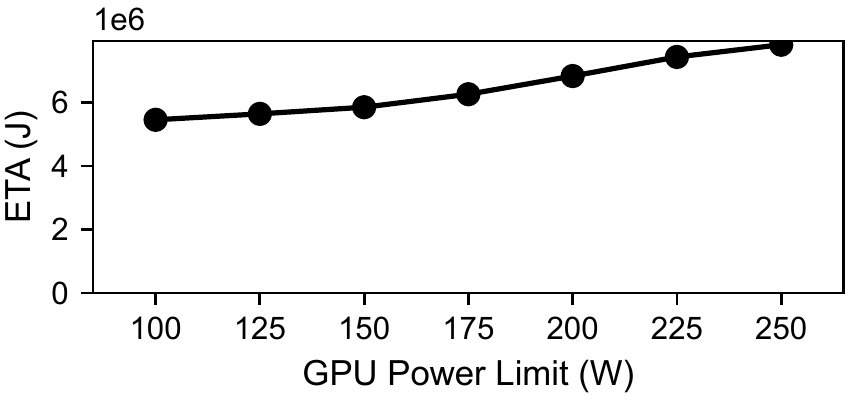}
	}
	\hfil
	\subfloat[][BERT (QA)]{
		\includegraphics[width=0.45\linewidth]{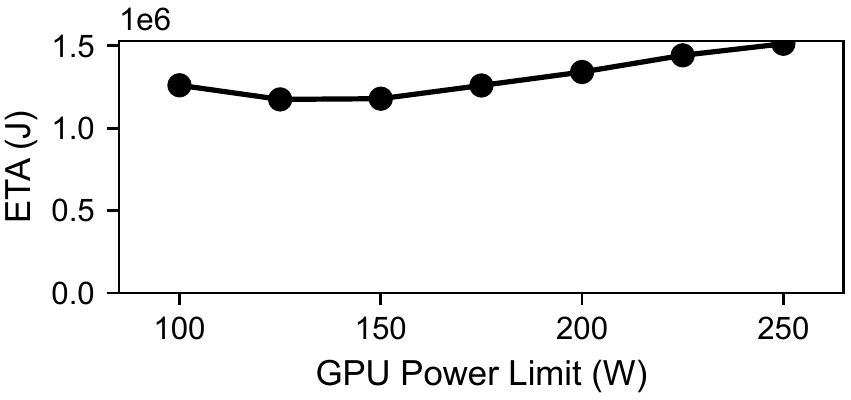} 
	}
	\hfil
	\subfloat[][BERT (SA)]{
		\includegraphics[width=0.45\linewidth]{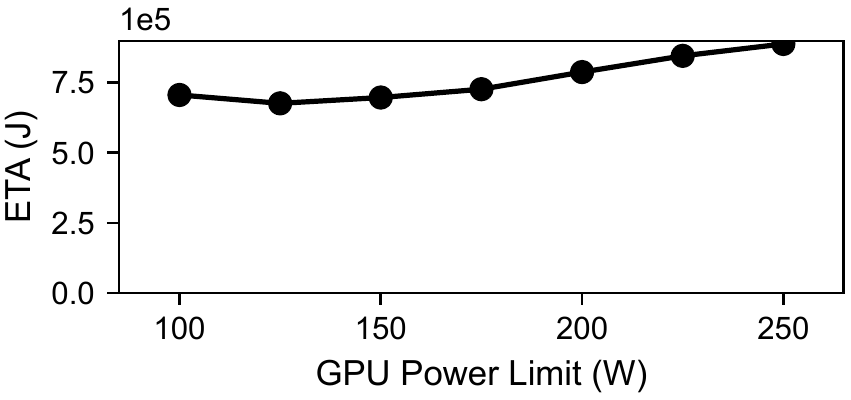}
	}
	\hfil
	\subfloat[][ResNet-50]{
		\includegraphics[width=0.45\linewidth]{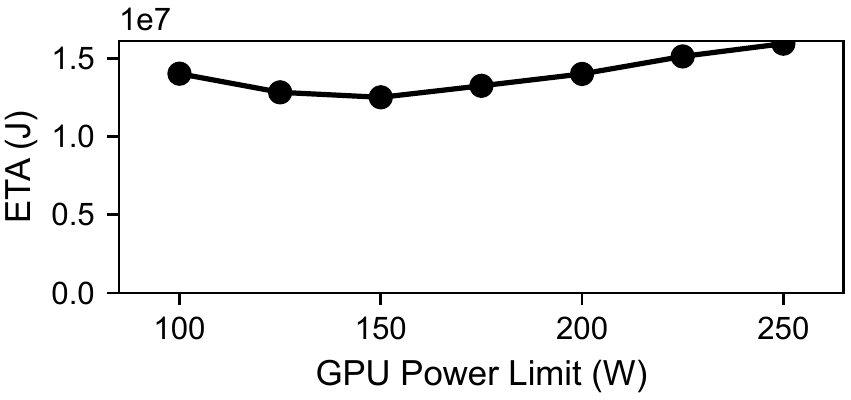}
	}
	\hfil
	\subfloat[][ShuffleNet V2]{
		\includegraphics[width=0.45\linewidth]{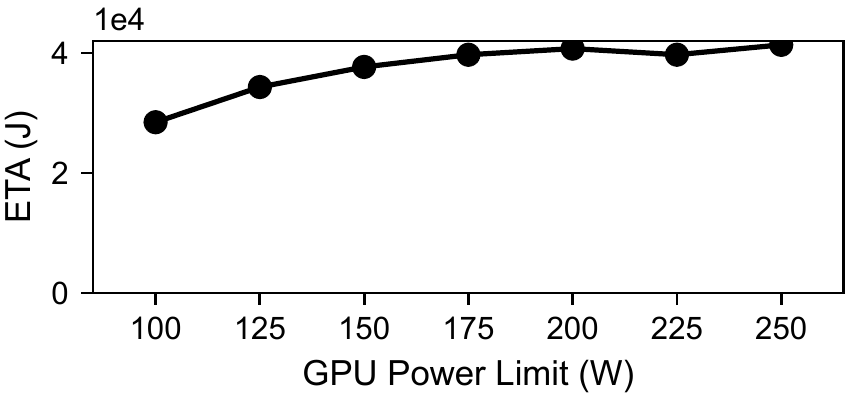}
	}
	\hfil
	\subfloat[][NeuMF]{
		\includegraphics[width=0.45\linewidth]{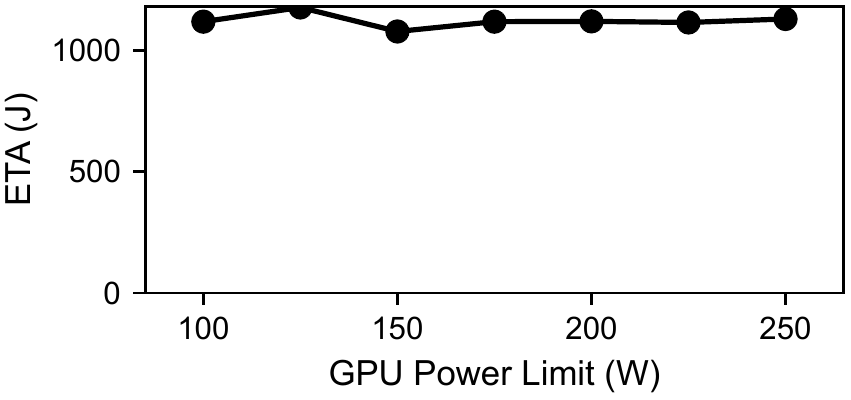}
	}
	\hfil
	
	\caption{ETA w.r.t GPU power limit of different DNN training workload. Measured on an NVIDIA V100 GPU.}
	\label{fig:appendix-pl-eta}
\end{figure}

Figures~\ref{fig:appendix-bs-eta} and~\ref{fig:appendix-pl-eta} respectively show the ETA value when batch size and power limit are swept.
Especially note the convexity of all BS-ETA curves, which justifies the design of our pruning exploration algorithm.


\section{Cumulative Regret of All Workloads}\label{sec:appendix-regret}

\begin{figure}[!t]
	\centering
	\hfil
	\subfloat[][DeepSpeech2]{
		\includegraphics[width=0.45\linewidth]{Figures/eval/regret-librispeech-eta-0.5.pdf}
	}
	\hfil
	\subfloat[][BERT (QA)]{
		\includegraphics[width=0.45\linewidth]{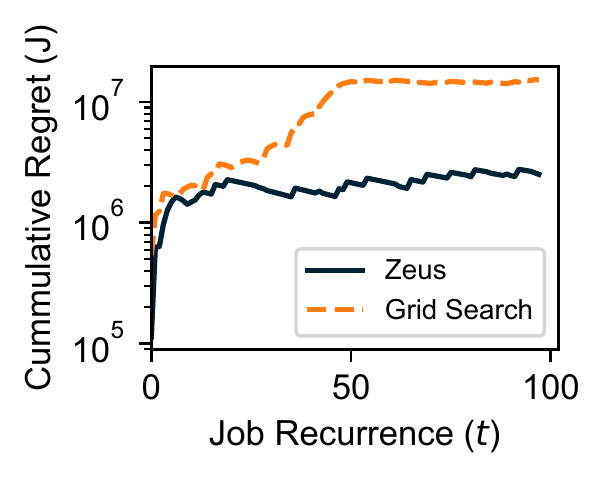} 
	}
	\hfil
	\subfloat[][BERT (SA)]{
		\includegraphics[width=0.45\linewidth]{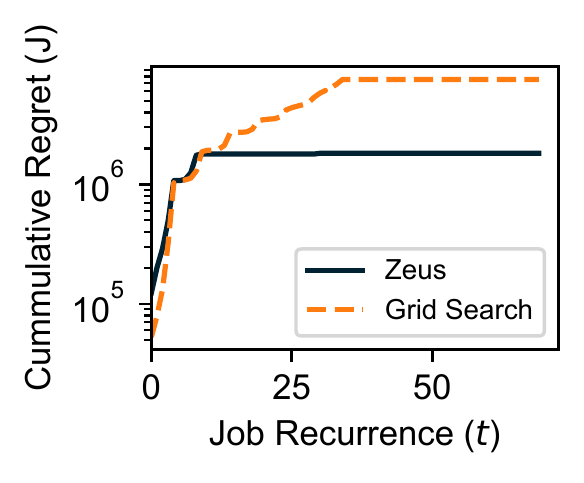}
	}
	\hfil	
	\subfloat[][ResNet-50]{
		\includegraphics[width=0.45\linewidth]{Figures/eval/regret-imagenet-eta-0.5.pdf}
	}
	\hfil
	\subfloat[][ShuffleNet V2]{
		\includegraphics[width=0.45\linewidth]{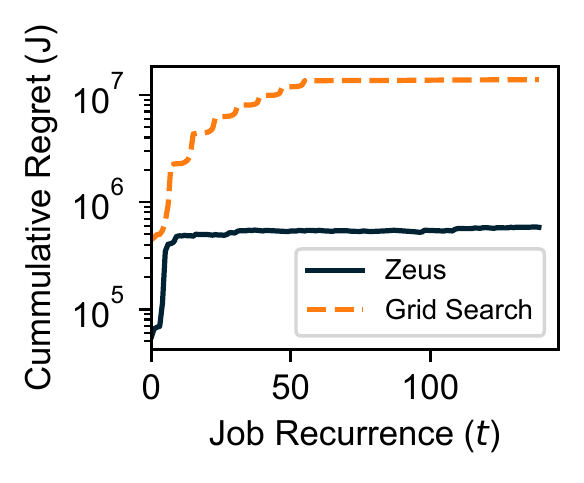}
	}
	\hfil
	\subfloat[][NeuMF]{
		\includegraphics[width=0.45\linewidth]{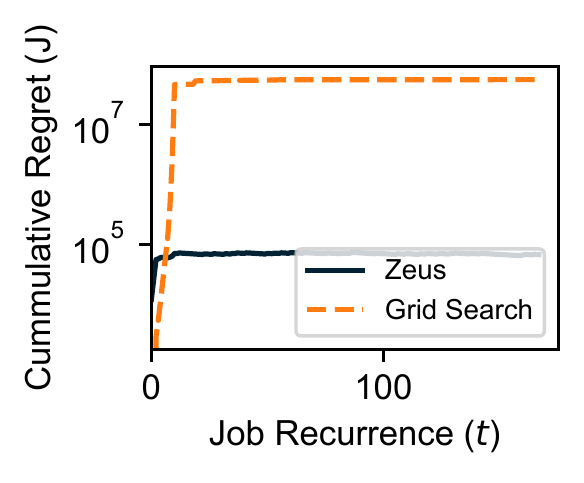}
	}
	\hfil
	\caption{Cumulative regret of Zeus vs. Grid Search across all workloads.}
	\label{fig:appendix-regret}
\end{figure}

Figure~\ref{fig:appendix-regret} shows the cumulative regret of {\name} and Grid Search over job recurrences for all six workloads.
In general, {\name} converges to a better knob than Grid Search while being faster.

\section{Search Paths for All Workloads}
\label{sec:appendix-searchpath}

\begin{figure}[!htbp]
	\centering
	\hfil
	\subfloat[][DeepSpeech2]{
		\includegraphics[width=0.45\linewidth]{Figures/eval/cov-heat-zeus-librispeech-eta-0.5.pdf}
	}
	\hfil
	\subfloat[][BERT (QA)]{
		\includegraphics[width=0.45\linewidth]{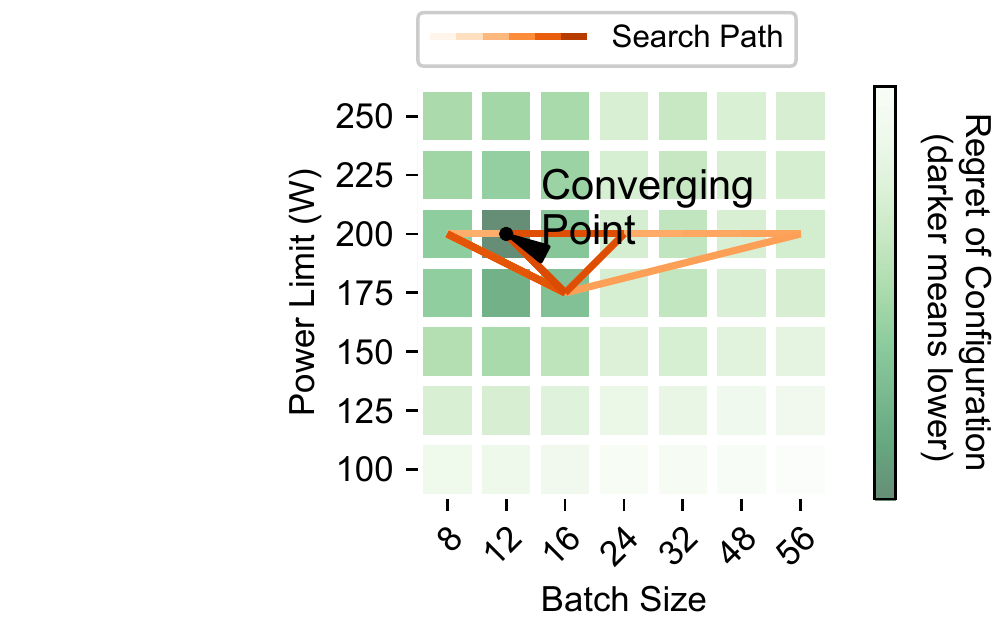} 
	}
	\hfil
	\subfloat[][BERT (SA)]{
		\includegraphics[width=0.45\linewidth]{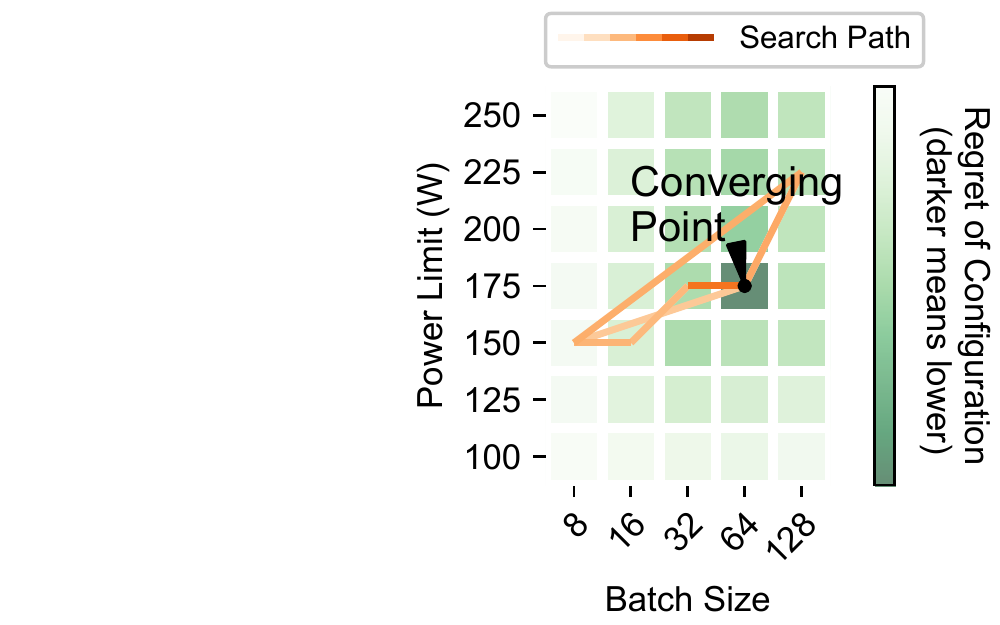}
	}
	\hfil
	\subfloat[][ResNet-50]{
		\includegraphics[width=0.45\linewidth]{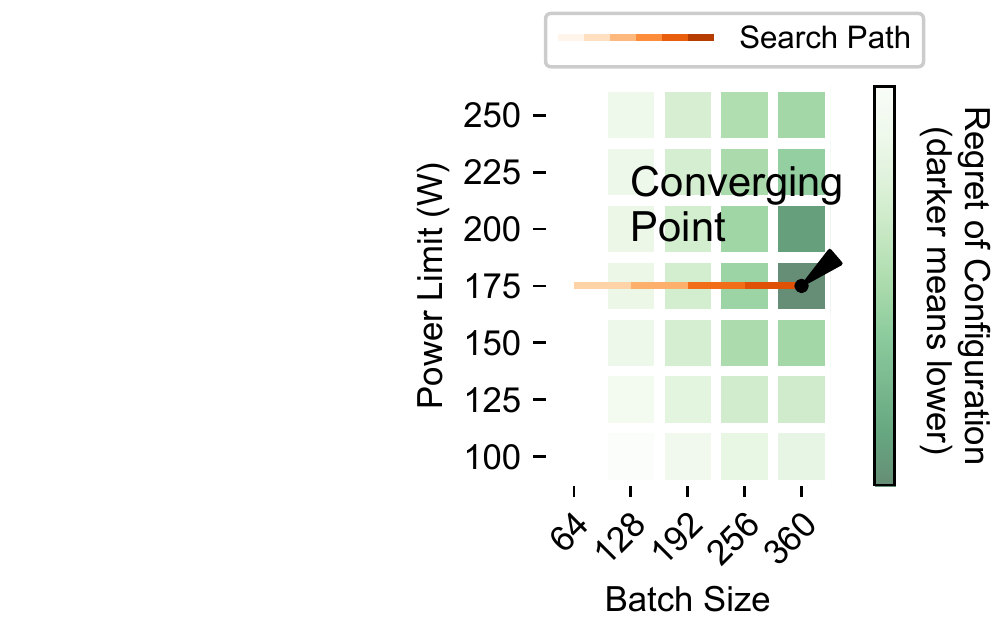}
	}
	\hfil
	\subfloat[][ShuffleNet V2]{
		\includegraphics[width=0.45\linewidth]{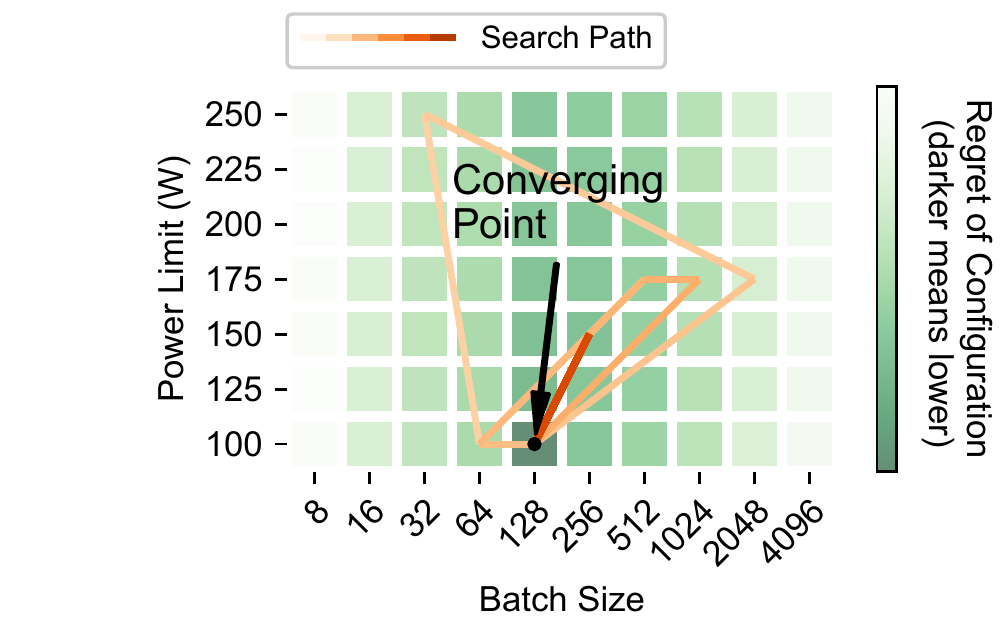}
	}
	\hfil
	\subfloat[][NeuMF]{
		\includegraphics[width=0.45\linewidth]{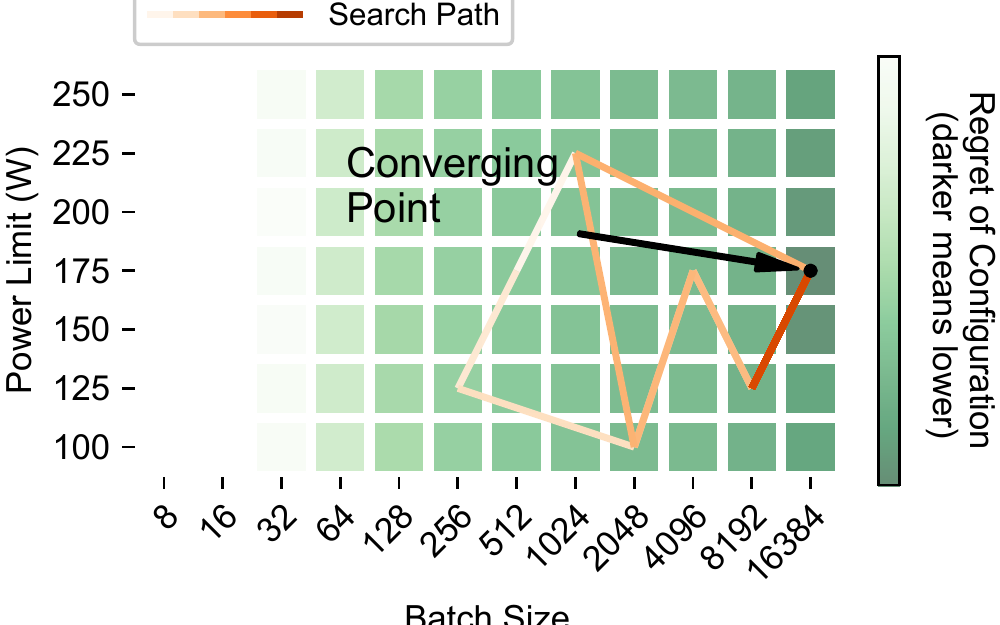}
	}
	\hfil
	\caption{Search path of Zeus across all workloads.}
	\label{fig:appendix-searchpath-zeus}
\end{figure}

\begin{figure}[!htbp]
	\centering
	\hfil
	\subfloat[][DeepSpeech2]{
		\includegraphics[width=0.45\linewidth]{Figures/eval/cov-heat-gs-librispeech-eta-0.5.pdf}
	}
	\hfil
	\subfloat[][BERT (QA)]{
		\includegraphics[width=0.45\linewidth]{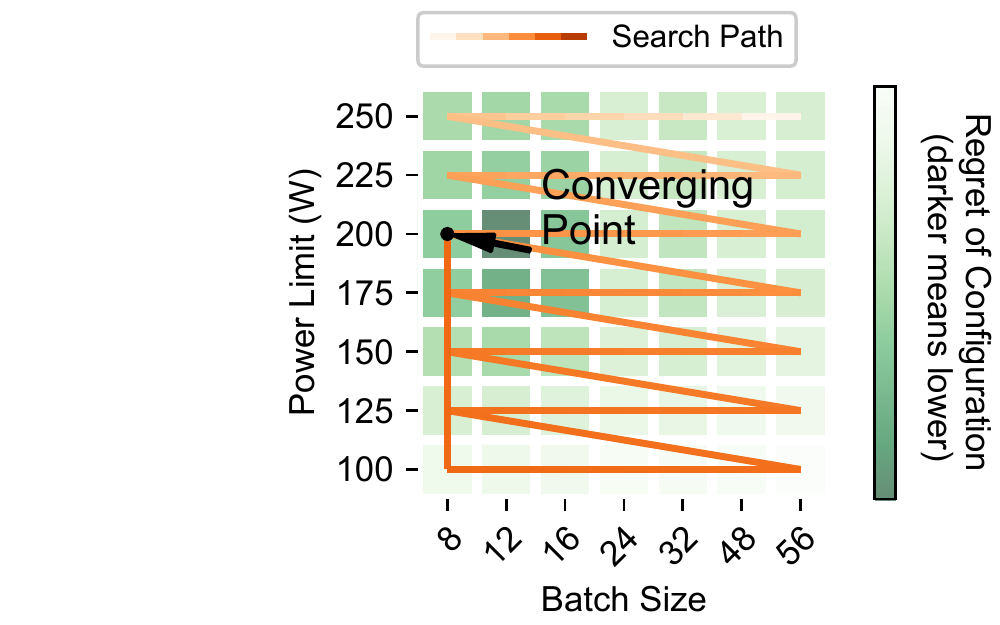} 
	}
	\hfil
	\subfloat[][BERT (SA)]{
		\includegraphics[width=0.45\linewidth]{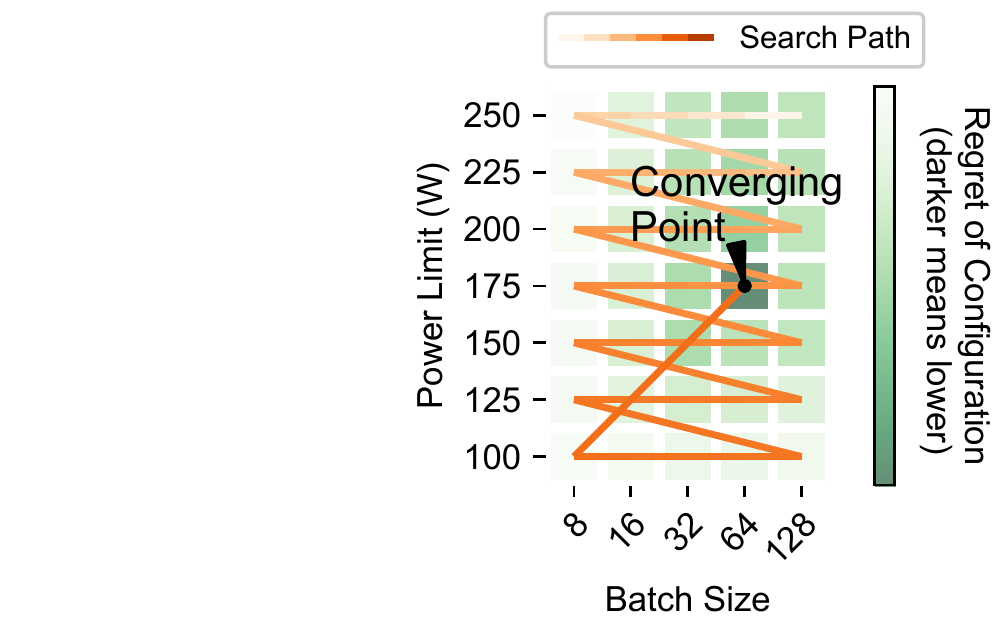}
	}
	\hfil
	\subfloat[][ResNet-50]{
		\includegraphics[width=0.45\linewidth]{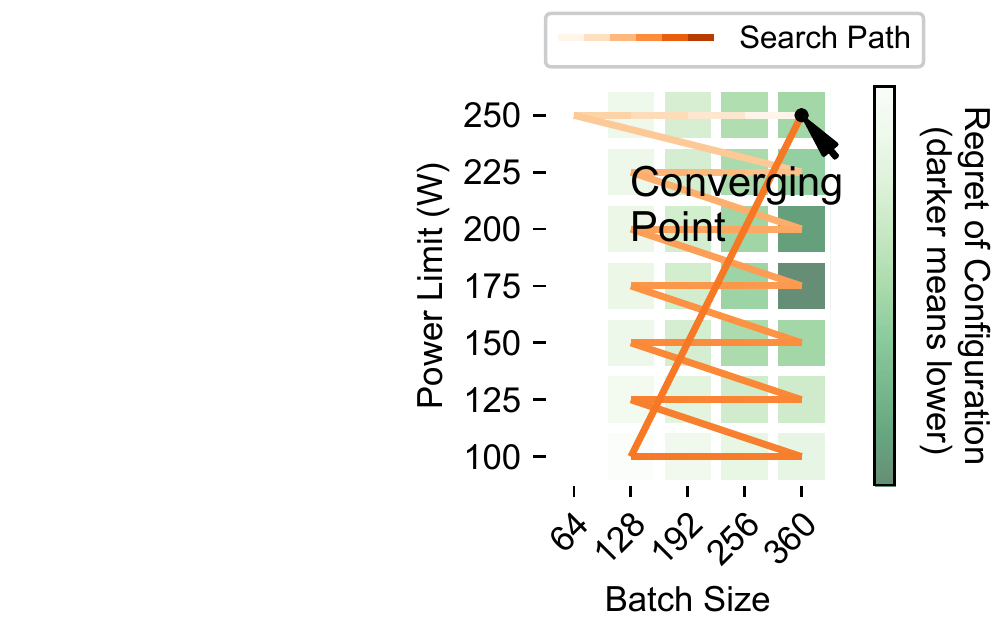}
	}
	\hfil
	\subfloat[][ShuffleNet V2]{
		\includegraphics[width=0.45\linewidth]{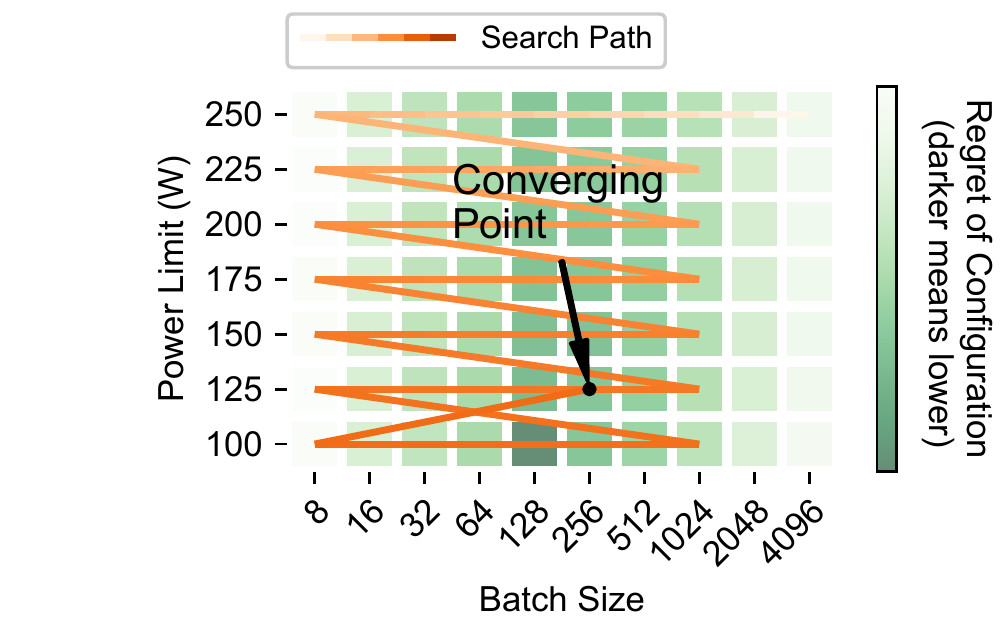}
	}
	\hfil
	\subfloat[][NeuMF]{
		\includegraphics[width=0.45\linewidth]{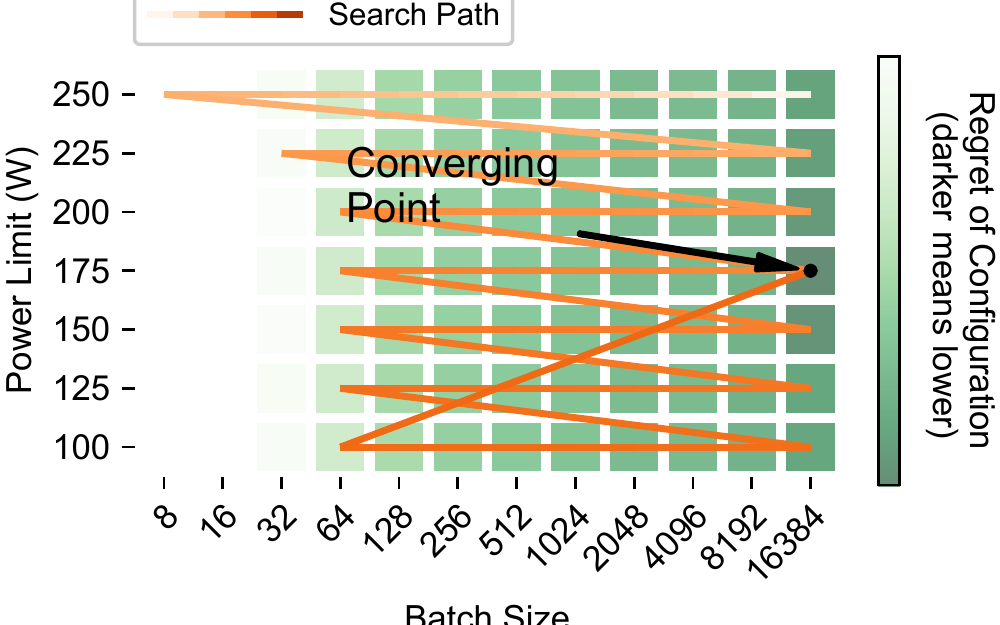}
	}
	\hfil
	\caption{Search path of Grid Search across all workloads.}
	\label{fig:appendix-searchpath-gs}
\end{figure}

Figures~\ref{fig:appendix-searchpath-zeus} and~\ref{fig:appendix-searchpath-gs} respectively show the search path of Zeus and Grid Search in the 2D configuration space.
Thanks to the decoupling of batch size and power limit, {\name} is able to more efficiently navigate the search space and converge to a knob while consuming less energy and time during exploration.

\section{Additional Sensitivity Analysis}
\label{sec:appendix-sensitivity}

\begin{figure}[!t]
	\centering
	\subfloat[][ETA]{
		\includegraphics[width=0.6\linewidth]{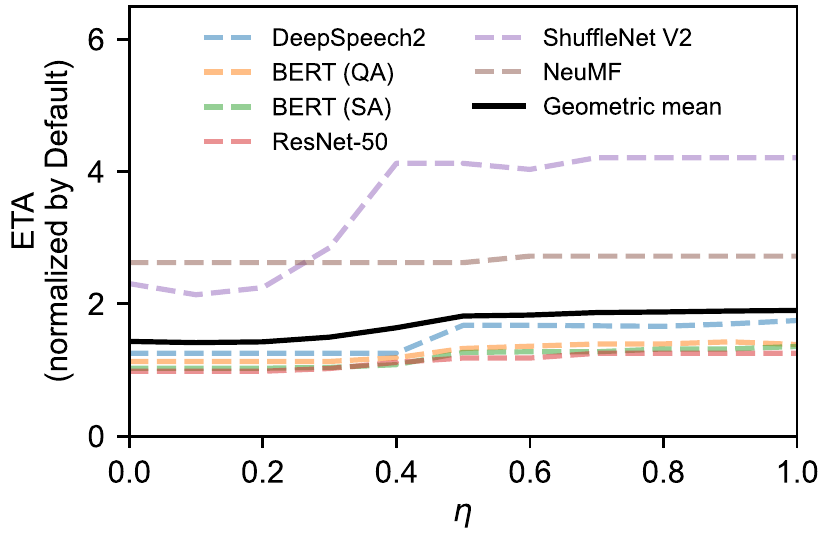}
	}
	\hfil
	\subfloat[][TTA]{
		\includegraphics[width=0.6\linewidth]{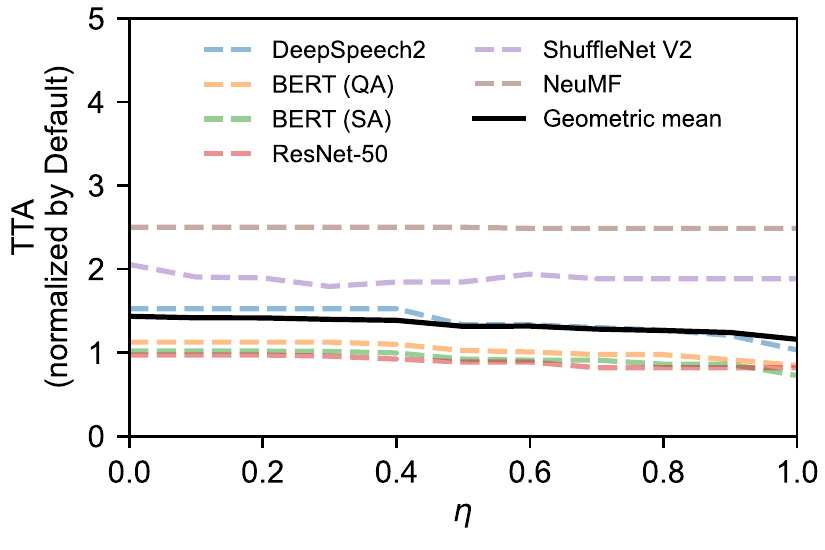}
	}
	\caption{Impact of priority knob $\eta$ on ETA and TTA.}
	\label{fig:appendix-sensitivity-etaknob}
\end{figure}

Figure~\ref{fig:appendix-sensitivity-etaknob} compares both the energy consumption and training time for {\name} against Default.
We also calculate and plot the geometric mean across all jobs.
The result shows that with higher $\eta$, {\name} prioritizes reducing energy consumption over time, leading to higher improvement factor of energy, and vice versa.



\section{Performance of {\name} on All GPUs}\label{sec:appendix-trace-foi-gpus}

\begin{figure}[!t]
	\centering
	\subfloat[][Energy Consumption (V100)]{
		\includegraphics[width=0.46\linewidth]{Figures/eval/cost-e-foi-v100-eta-0.5.pdf}
	}
	\hfil
	\subfloat[][Training Time (V100)]{
		\includegraphics[width=0.46\linewidth]{Figures/eval/cost-t-foi-v100-eta-0.5.pdf}
	} \\
	\subfloat[][Energy Consumption (A40)]{
		\includegraphics[width=0.46\linewidth]{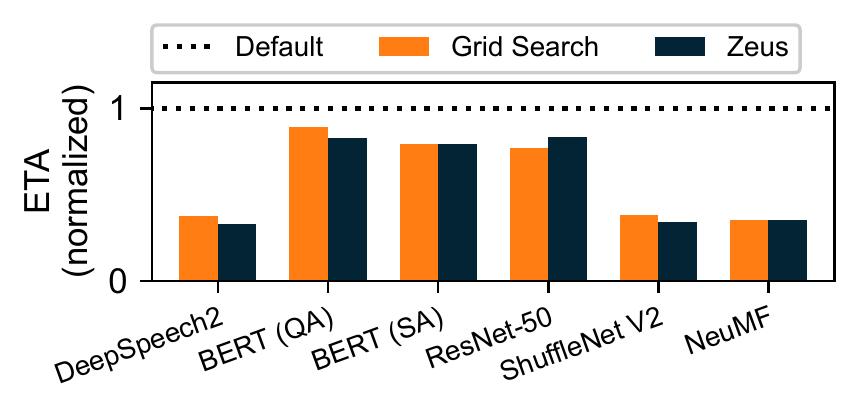}
	}
	\hfil
	\subfloat[][Training Time (A40)]{
		\includegraphics[width=0.46\linewidth]{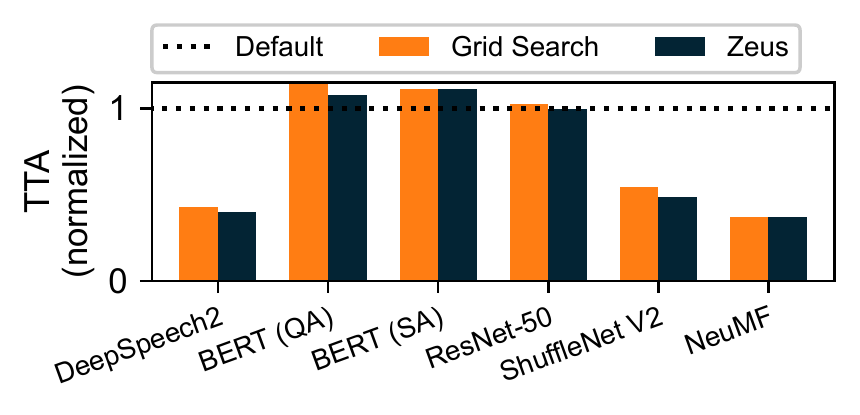}
	} \\
	\subfloat[][Energy Consumption (RTX6000)]{
		\includegraphics[width=0.46\linewidth]{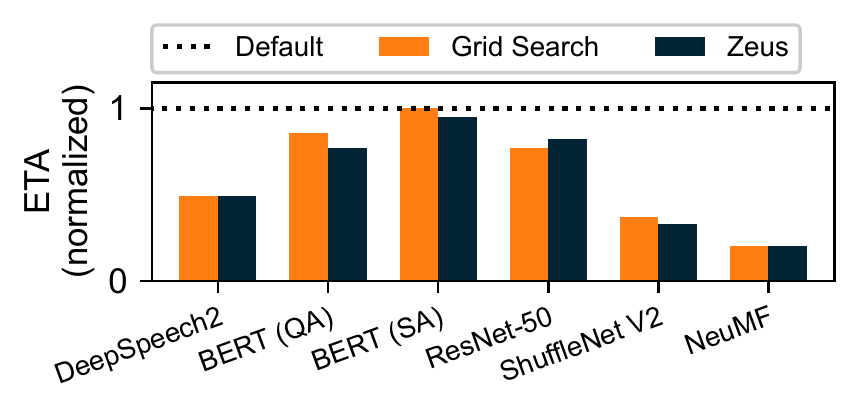}
	}
	\hfil
	\subfloat[][Training Time (RTX6000)]{
		\includegraphics[width=0.46\linewidth]{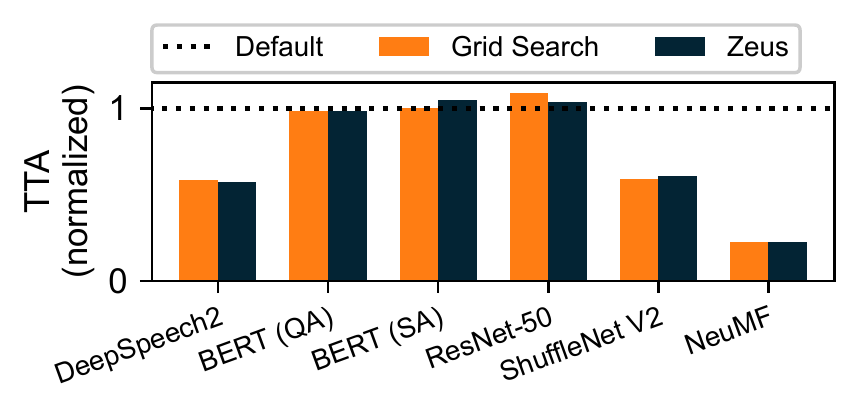}
	} \\
	\subfloat[][Energy Consumption (P100)]{
		\includegraphics[width=0.46\linewidth]{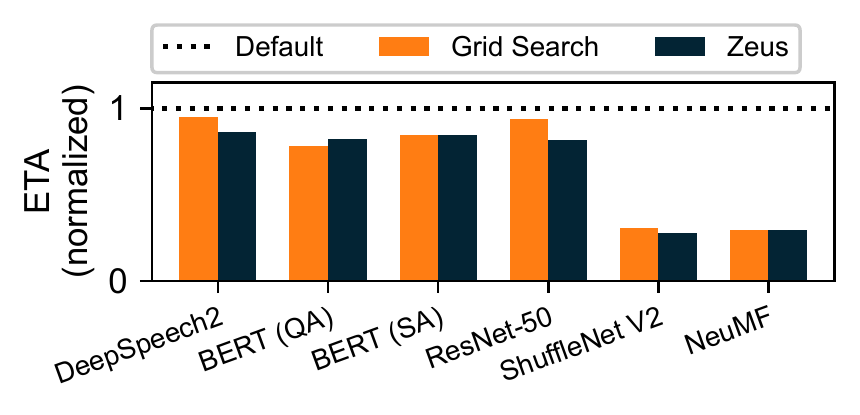}
	}
	\hfil
	\subfloat[][Training Time (P100)]{
		\includegraphics[width=0.46\linewidth]{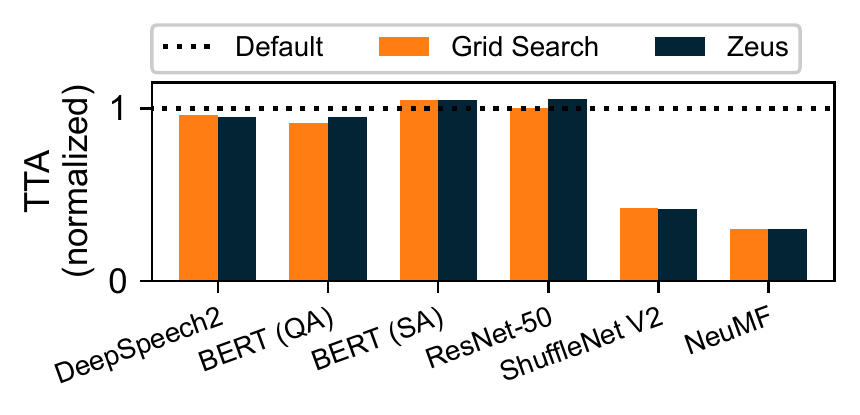}
	} 
	\caption{Energy and time consumption of DNN training, normalized against Default for DNN training. Results measured on (a) NVIDIA A40 GPU, (b) NVIDIA V100 GPU, (c) NVIDIA RTX6000 GPU and (d) NVIDIA P100 GPU.}
	\label{fig:appendix-trace-foi-gpus}
\end{figure}

Figure~\ref{fig:appendix-trace-foi-gpus} presents the energy and time consumption of all workloads on four different generations NVIDIA GPUs: Ampere (A40), Volta (V100), Turing (RTX6000), and Pascal (P100).
The overall trends hold for all GPUs.